%% file: main.tex
\documentclass{vldb}
\usepackage{amssymb}
\usepackage{amsmath}
\usepackage{graphicx}
\usepackage{color}
\usepackage{bbm}
\usepackage{balance}
\usepackage{booktabs}
\usepackage[para,flushleft]{threeparttable}
\usepackage{caption}
\usepackage{subcaption}
\usepackage{comment}
\usepackage{enumitem}
\usepackage{float}
\usepackage[noblocks]{authblk}
\newtheorem{definition}{Definition}[section]

\usepackage{lipsum}

\begin{document}

\title{Network Representation Learning: Consolidation and Renewed Bearing} 
\author[1] {Saket Gurukar \thanks{ Equal contribution. Rest of the authors are listed in the alphabetical order of their last names}}
\author[2, 4]{Priyesh Vijayan\textsuperscript{*}}
\author[2]{Aakash Srinivasan\textsuperscript{*}}
\author[1]{Goonmeet Bajaj}
\author[1]{Chen Cai}
\author[1]{Moniba Keymanesh}
\author[1]{Saravana Kumar}
\author[1]{Pranav Maneriker}
\author[3]{Anasua Mitra}
\author[1]{Vedang Patel}
\author[2, 4]{Balaraman Ravindran}
\author[1]{Srinivasan Parthasarathy}

\affil[1]{Computer Science and Engineering, The Ohio State University}
\affil[2]{Department of Computer Science and Engineering, IIT Madras}
\affil[3]{Department of Computer Science and Engineering, IIT Guwahati}
\affil[4]{Robert Bosch Centre for Data Sciences and AI, IIT Madras}
\maketitle

\begin{abstract}

Graphs are a natural abstraction for many problems where nodes represent entities and edges represent a relationship across entities. The abstraction can be explicit (e.g., transportation networks, social networks, foreign key relationships) or implicit (e.g., nearest neighbor problems).  An important area of research that has emerged over the last decade is the use of graphs as a vehicle for non-linear dimensionality reduction in a manner akin to previous efforts based on manifold learning with uses for downstream database processing (e.g., entity resolution and link prediction, outlier analysis), machine learning and visualization.  In this systematic yet comprehensive experimental survey, we benchmark several popular network representation learning methods operating on two key tasks: link prediction and node classification.

We examine the performance of 12 unsupervised embedding methods on 15 datasets. To the best of our knowledge, the scale of our study -- both in terms of the number of methods and number of datasets -- is the largest to date. Our benchmarking study in as far as possible uses the original codes provided by the original authors.

    Our results reveal several key insights about work-to-date in this space. First, we find that certain baseline methods (task-specific heuristics, as well as classic manifold methods) that have often been dismissed or are not considered by previous efforts can compete on certain types of datasets if they are tuned appropriately. Second, we find that recent methods based on matrix factorization offer a small but relatively consistent advantage over alternative methods (e.g., random-walk based methods) from a qualitative standpoint. Specifically, we find that MNMF, a community preserving embedding method, is the most competitive method for the link prediction task. While NetMF is the most competitive baseline for node classification. Third, no single method completely outperforms other embedding methods on both node classification and link prediction tasks. We also present several drill-down analysis that reveals settings under which certain algorithms perform well (e.g., the role of neighborhood context on performance; dataset characteristics that influence performance) -- guiding the end-user.
\end{abstract}
\section{Introduction}

\input{introduction.tex}

\section{Notations and Definitions}
\input{notations_and_definitions.tex}

\section{Network Embedding methods} 
\label{section:models}
\input{methods.tex}

\section{Datasets}
\input{datasets.tex}

\section{Experimental setup}
\input{experimental_setup.tex}

\section{Experimental Results}

\input{experimental_results.tex}



\section{Discussions and Conclusions}
\input{conclusion.tex}

\input{tables.tex}

\clearpage
\bibliographystyle{abbrv}
\bibliography{vldb} 

\end{document}

%% file: introduction.tex
Graphs are effective in multiple disparate domains to model, query and mine relational data. Examples abound ranging from the use of nearest neighbor graphs in database systems ~\cite{eppstein1997nearest, zhao2009anomaly} and machine learning ~\cite{bhattacharyya2013network, nickel2016review} to the analysis of biological networks ~\cite{hamilton2017inductive, benson2016higher} and from social network analysis~\cite{gu2017co,zhang2017olak} to the analysis of transportation networks\cite{cetinkaya2015multilevel}.  ML-enhanced data structures and algorithms such as learned indexes \cite{kraska2018case} have recently shown promising results in database systems. An active area in ML research -- network representation learning --  has potential in multiple applications related to the downstream database processing tasks such as outlier analysis \cite{zhao2009anomaly, liang2018semi}, entity resolution \cite{cohen2002learning, getoor2012entity}, link prediction \cite{liben2007link, perozzi2014deepwalk} and visualization \cite{ eckardt2010method, ribeiro2017struc2vec}.
However, a plethora of new network representation learning methods has been proposed recently \cite{cai2018comprehensive, hamilton2017representation}. Given the wide range of methods proposed, it is often tough for a practitioner to determine or understand which of these methods they should consider adopting for a particular task on a particular dataset. 
Part of the challenge is the lack of a standard evaluation benchmark and a thorough independent understanding of the strengths and weaknesses of each method for the particular task on hand. The challenges are daunting and can be summarized as follows:

\smallskip
\noindent {\bf Lack of Standard Assessment Protocol:} First, there is no standard to evaluate the quality of generated embeddings. The efficacy among embedding methods is often evaluated based on downstream machine learning tasks. As a result, the superiority of one embedding method over another \textbf{hinges} on the performance in a downstream machine learning task. With the lack of a standard evaluation protocol for these downstream tasks, the results reported by different research articles are often inconsistent. 
As a specific example, Node2vec \cite{grover2016node2vec} report the node classification performance of Deepwalk on Blogcatalog dataset -- for multi-label classification with a train-test split of 50:50  -- as 21.1\% Macro-f1, whereas the Deepwalk paper \cite{perozzi2014deepwalk} reports Deepwalks' performance as 27.3\% Macro-f1.  

\smallskip
 \noindent {\bf Tuning Comparative Strawman:} Second, a new method almost always compares its performance against a subset of other methods, and on a subset of tasks and datasets previously evaluated. In many cases, while great care is taken to tune the new method (via careful hyper-parameter tuning) -- the same care is often not taken when evaluating baselines. For example, in our experiments on Blogcatalog, we find that with a train-test split of 50:50 the Laplacian Eigenmaps method \cite{belkin2003laplacian} without Grid-Search achieves a Macro-f1 score of \textbf{3.9\%} (similar to what was reported in \cite{goyal2018graph, grover2016node2vec}). However, with tuning the hyper-parameters of logistic regression, we find that the Laplacian Eigenmaps method achieves a Macro-F1 of \textbf{29.2\%}. 
Importantly, while logistic regression is commonly used to evaluate the quality of node embeddings in such methods, Grid-Search over logistic regression parameters is rarely conducted or reported.  Additionally, reported results are rarely averaged over multiple shuffles to reduce any bias or patterns in the training data\footnote{This is our observation based on the evaluation scripts publically shared by multiple authors.}. In short, a lack of consistency in evaluation inhibits our understanding of the scientific advances in this arena, discussed next.

\smallskip
\noindent {\bf  Standard Benchmark:}
Third, there is no agreed list of datasets that are used consistently in the literature. A new embedding method evaluates their method on selected datasets with a suitable node classification/link prediction setup. For instance, few methods report node classification performance for the baselines with the train-test split of 10:90 while few methods report the same with the train-test split of 50:50. As a result, the comparison across embedding methods is often unclear. Additionally, there are no clear guidelines on whether the proposed embedding methodology favors a certain type of dataset characteristic (e.g., sparsity).

\smallskip
\noindent {\bf  Task Specific Baselines:} Fourth, for many tasks such as node classification and link prediction there is a rich pre-existing literature \cite{bhagat2011node, lu2011link} focused on such tasks (that do not explicitly rely on node embedding methodology as a preprocessing step). Few, if any, of the prior art in network representation learning consider such baselines --  often such methods compare performance on downstream ML tasks against other node embedding methods. 
In our experiments, we find that a curated feature vector based on heuristics can achieve a similar competitive AUROC score on many of the datasets for the link prediction task. 

To summarize, there is a clear and pressing need for a comprehensive and careful benchmarking of such methods which is the focus of this study. To address the aforementioned issues in the network embedding literature, we perform an experimental study of 12 promising network embeddings methods on 15 diverse datasets. The selected embedding methods are unsupervised techniques to generate the node embeddings of a graph. 
Our goal is to perform a uniform, principled comparison of these methods on a variety of different datasets and across two key tasks -- link prediction and node classification.



Specific findings of our experimental survey that we wish to highlight are that:

\begin{enumerate}[leftmargin=*]
    \item For the link-prediction task we find that  MNMF \cite{wang2017community}, a community preserving embedding method, offers a compelling advantage when it scales to a particular dataset. Other more scalable alternatives such as Verse and LINE also perform well on most datasets. The heuristic approach we present for link prediction competes exceptionally well on all types of datasets surveyed for this task. 
    \item For the node classification task,  NetMF \cite{qiu2018network} when it scales to the particular dataset, offers a small but consistent performance advantage. 
    We find that for node-classification task, the task-specific heuristic methodology we compare with works well when operating on datasets with fewer labels -- in such scenarios, it competes well with a majority of the methods surveyed, whereas, some recent methods proposed fare much worse.
    \item We also drill down to study the impact of context embeddings on the link prediction and node classification tasks (and find some methods impervious to the use of context -- where for others it helps significantly). We also examine two common ways in which link prediction strategies are evaluated (explicitly through a classifier, or implicitly through vector dot-product ranking). We find that there is a clear separation in performance when using these alternative strategies.

\end{enumerate}

%% file: notations_and_definitions.tex
We denote the input graph as $\mathcal{G} = (V, E)$ where $V$ and $E$ denote the set of nodes and edges of the graph, $\mathcal{G}$. The notations used in this work are listed in Table \ref{notations}. In this study, we consider both directed as well as undirected graphs along with weighted as well as unweighted graphs. We evaluate the embedding methods on non-attributed, homogeneous graphs.

\begin{table}[ht]
\begin{center}
\begin{tabular}{l|l}
\hline
Symbol & Meaning \\
\hline
$\mathcal{G}$ & Input graph \\
$V$ & Nodes \\
$E$ & Edges \\
$n$ & Number of nodes, $|V|$ \\
$A$ & Adjacency  matrix. $A \in \mathbb{R}^{n*n} $\\
$D$ & Degree Matrix of Graph. \\ & $D_{i,i} =\sum_{j=0}^{n} A_{i,j} $ and $D_{i,j}=0$ where $ i \ne j $\\
$I$ & Identity Matrix \\
$\Phi(u)$ & Node embedding of node $u$ \\
$\psi(u)$ & Context embedding of node $u$ \\
$U, V$ & Node and context embedding matrix of size  $\mathbb{R}^{n*d}$ \\
$vol(G)$ & Sum of weights of all edges \\
$S$ & Graph Similarity matrix \\
$\sigma(x)$ & a non-linear function such as Sigmoid function \\ 
$\lambda$ & Number of negative samples \\
$P$ & $D^{-1}A$ \\
$L_{rw}$ &  $L_{rw} = I - D^{-1}A$ \\
$L_{sym}$ &  $L_{sym} = I - D^{-1/2}AD^{-1/2}$ \\
\hline
\end{tabular}
\end{center}
\caption{The notations symbols used in this study and the terms associated with those symbols.}
\label{notations}
\end{table}

\begin{definition}{Network Embedding:}
Given a Graph, $\mathcal{G} = (V, E)$ and an embedding dimension,  $d$ where $d \ll |V|$, the goal of a network embedding method is to learn a $d$-dimensional representation of the graph,  $\mathcal{G}$ such that similarity in graph space approximates closeness in $d$-dimensional space.
\end{definition}

%% file: methods.tex
\label{network_methods}
In this section, we give a summary of the network embedding methods evaluated in our work. Herein, for each models along with their description, we also provide additional experimental details for reproducibility. 



\begin{enumerate}

\item \textbf{Laplacian Eigenmaps \cite{belkin2003laplacian}}: Laplacian Eigenmaps generates a $d$-dimensional embedding of the graph using the smallest $d$ eigenvectors of Laplacian matrix $L = D-A$.
\begin{equation*}
\begin{aligned}& \underset{U}{\text{minimize}}& & trace(U^TLU) \\
& \text{subject to}& & U^TDU = I
\end{aligned}
\end{equation*}
$U$ is generated embedding matrix $\mathbb{R}^{|V| * d}$. The above equation can be reduced to simple minimization of L2 distance for adjacent nodes - $\Sigma_{i,j}||u^{(i)}-u^{(j)}||^2A_{ij}$. 
Laplacian Eigenmaps levers the first order information for generating the embeddings.

\textbf{Reproducibility notes}: We search for following  hyperparameters: Embedding dimension = [64, 128, 256]. On the datasets with $>$1M nodes, Laplacian Eigenmaps did not scale for embedding dimension 128, 256. 

\item \textbf{DeepWalk \cite{perozzi2014deepwalk}}: DeepWalk is a random walk based network embedding method which uses truncated random walks and levers local information from these generated walks to learn similar latent representations. 
DeepWalk draws inspiration from Skip-gram model in Word2vec \cite{MikolovWord2vec} by treating random walks as sequences and optimizing following objective function:
\begin{equation}
minimize_{\Phi} \; - \log Pr(\{v_{i-w},\cdots, v_{i+w} \} \backslash  v_i \; | \; \Phi(v_i))
\end{equation}
where  $v_{i}$ is target node while $\{v_{i-w},\cdots, v_{i+w} \} \backslash  v_i$ are the context nodes. $\Phi(v_{i})$ denotes the embedding of the node $v_{i}$. Since the objective function is expensive to compute for large graphs, it is approximated using Hierarchical Softmax\cite{mnih2009scalable}.

\textbf{Reproducibility notes}: We search for following  hyperparameters: Walk length = [5, 20, 40], Number of walks = [20, 40, 80], Window size = [2, 4, 10], Embedding dimension = [64, 128, 256]. In the case of directed graphs, we observe lower performance in node classification and link prediction task. In order to have a fair comparison with other methods, we treat directed graphs as undirected for DeepWalk.

\item \textbf{Node2Vec \cite{grover2016node2vec}}: Node2Vec is a biased random walk based network embedding method which allows the random walk to be more flexible in exploring the graph neighborhoods. The flexibility of the random walk is achieved by interpolating between Breadth-first traversal and Depth-first traversal. The objective function is again based on Skip-gram model -based on Word2vec \cite{MikolovWord2vec}, and since the objective function is expensive to compute for large graphs, it is approximated by negative sampling \cite{MikolovWord2vec}.

\textbf{Reproducibility notes}: We search for following  hyperparameters: Walk length = [10, 20, 40], Number of walks = 80. Window Size = 10, $p$ and $q$ = [0.25, 1, 2, 4], Embedding dimension = [64, 128, 256].  In case of directed graphs, we observe lower performance in node classification and link prediciton task. In order to have a fair comparison with other methods, we treat directed graphs as undirected for Node2Vec.

\item \textbf{GraRep \cite{cao2015grarep}}: GraRep is a matrix factorization based network embedding method which captures the global structural information of the graph while learning node embeddings. The authors observe that the existing Skip-gram based models project all the $k$-step relational information into a common subspace and then, argue the importance of preserving different $k$-step relational information in separate subspaces. The loss function to preserve the $k$-step relationship between node $w$ and $c$ is proposed as: 
\begin{equation}
\begin{aligned}
L_k(u, v) = & A_{u, v}^{k}\cdot\log{\sigma(\Phi(u) \cdot \Phi(v))} \\ + & \frac{\lambda}{|V|}\sum\limits_{v' \in V, v' \ne v}{A_{u, v'}^k \cdot \log{\sigma(-\Phi(u)\cdot\Phi(v'))}}
\end{aligned}
\end{equation}

where $v'$ refers to the negative node at $k$-th step for node $u$ 
(see Table 1 for additional notation, e.g. $\lambda$). The above loss function in closed form results in log-transformed, probabilistic adjacency matrix which is factorized with SVD for generating each $k$-step representation. The final node representation is generated by concatenation of all the $k$-step representations.

\textbf{Reproducibility notes}: We search for following  hyperparameters: $k$ from 1 to 6, Embedding dimension = [64, 128, 256]. On the datasets with $>$2M edges, due to scalability issue, we searched for $k$ from 1 to 2 and Embedding dimension = [64, 128].

\item \textbf{NetMF \cite{qiu2018network}}: NetMF is a matrix factorization based network embedding method. NetMF presents theoretical proofs for their claim that Skip-gram models with negative sampling are implicitly approximating and factorizing appropriate matrices constructed with the help of graph Laplacians. The objective matrix based for NetMF on small context window T is given by (see Table 1 for notation):
\begin{equation}
    \small{
    \log \left(\operatorname{vol}(G)\left(\frac{1}{T} \sum_{r=1}^{T}\left(D^{-1} A\right)^{r}\right) D^{-1}\right)-\log b
    }
\end{equation}
where $\operatorname{vol}(G)$ refers to sum of all edge weights and $b$ corresponds to number of negative samples in skip-gram model.  
NetMF factorizes the above closed form DeepWalk matrix with SVD in order to generate node embedding and provides two algorithms for small context window and large context window.

\textbf{Reproducibility notes}: We search for following  hyperparameters: $T$ = [1, 10], Negative samples $\lambda$ = [1, 2, 3], Rank $H$ for large context window  = [128, 256, 512], Embedding dimension = [64, 128, 256].

\item \textbf{M-NMF \cite{wang2017community}}: M-NMF is a matrix factorization based network embedding method which generates node embeddings that preserves the microscopic information in form of first-order and second-order proximities among nodes and the generated embeddings also preserve mesoscopic information in form of community structure. The objective function for M-NMF is given as 
\begin{equation} 
\begin{aligned}
 O = \min_{ U,V C, H \geq 0} & {\lVert S-VU^{T} \rVert}^2 + \alpha * {\lVert H-UC^{T} \rVert}^2_F \\ & - \beta {tr}(H^TBH) + \zeta {\lVert HH^T-I \rVert}^2_F
\end{aligned}
\end{equation}
where $H$ is the binary community membership matrix, $C$ is the latent representations of communities and $B$ is the modularity matrix obtained from the adjacency matrix, $A$ (see Table \ref{notations} for rest of the notations). Overall, M-NMF discovers communities through modularity constraints. The node embeddings generated with the help of microscopic information and community embeddings are then, jointly optimized by assuming consensus relationship between both node and community embeddings.

\textbf{Reproducibility notes}: We search for following  hyperparameters: $\alpha$ = [0.1, 1.0, 10.0], $\beta$ = [0.1, 1.0, 10.0], Embedding dimension = [64, 128, 256].

\item \textbf{HOPE \cite{ou2016asymmetric}}: HOPE is a matrix factorization based network embedding method which generates node embeddings that preserve asymmetric transitivity of nodes in directed graphs. If there exists a directed edge from node $u$ to $w$ and $w$ to $v$, then  -- due to asymmetric transitivity property -- an edge from $u$ to $v$ is more likely to form than edge from $v$ to $u$. The objective function of HOPE is given as follows
\begin{equation}
    min \;  \| S - U^s(U^t)^T \|^2_2
\end{equation}
where $U^s$ and $U^t$ are the source and target embeddings. In order to preserve asymmetric transitivity of nodes, the proximity matrix $S$ is constructed using a similarity metric which respects the directionality of edges. The node embeddings are generated by factorizing the proximity matrix with generalized SVD \cite{paige1981towards}.

\textbf{Reproducibility notes}: We search for following  hyperparameters: The decay parameter $\beta$ = 0.5/$\alpha$, where $\alpha$ is spectral radius of the graph. Embedding dimension = [64, 128, 256].

\item \textbf{LINE \cite{tang2015line}}: LINE is an optimization-based network embedding method which optimizes an objective function that preserves both first and second order proximity among nodes in the embedding space. The objective function for first-order proximity is given as:
\begin{equation}
    O_1 = -\sum_{(u,v) \in E} A_{u,v} \; log \; \sigma(\Phi(u). \Phi(v))
\end{equation}
The objective function to preserve the second order proximity is given as:
\begin{equation}
    O_2 = -\sum_{(u,v) \in E} A_{u,v}  \; log \; \frac{exp(\Phi(u). \psi(v))}{\sum_{v' \in V, \substack{v' \ne v}} exp(\Phi(u). \psi(v'))}
\end{equation}

where $\psi(u)$ represents context embedding of node $u$ (see Table \ref{notations} for rest of the notations). The first-order proximity corresponds to local proximity between nodes based on the presence of edges in the graph while the second-order proximity corresponds to global proximity between nodes based on shared neighborhoods of those nodes in the graph. Since the objective function is expensive to compute for large graphs, it is approximated by negative sampling \cite{MikolovWord2vec}.

\textbf{Reproducibility notes}: We search for following  hyperparameters: Number of samples = 10 billion, Embedding dimension $\in$ [64, 128, 256]. In the case of directed graphs, as suggested by the authors of LINE, we evaluate only second-order proximity.

\item \textbf{Verse \cite{tsitsulin2018verse}}: Verse is an optimization-based network embedding method which optimizes an objective function that minimizes the Kullback-Leibler (KL) divergence from the given similarity distribution in graph space to similarity distribution in embedding space (E). The objective function is given as follows:
\begin{equation}
    \sum_{v\in V} \; \text{KL} ( sim_G(v,.) \| sim_E(v,.))
\end{equation}
The similarity distribution in graph space could be constructed with help of Personalized PageRank\cite{page1999pagerank}, SimRank\cite{jeh2002simrank}, or Adjacency matrix\cite{tsitsulin2018verse}. Since the objective function is expensive to compute for large graphs, it is approximated by Noise Constrastive Estimation \cite{gutmann2010noise}. 

\textbf{Reproducibility notes}: We search for following  hyperparameters: PageRank damping factor $\alpha$ = [0.7, 0.85, 0.9], Negative samples = [3, 10], Embedding dimension = [64, 128, 256].

\item \textbf{SDNE \cite{wang2016structural}}: 
SDNE is a deep autoencoder based network embedding method which optimizes an objective function that preserves both first and second order proximity among nodes in the embedding space. The objective function of SDNE is given below
\begin{equation}
    \mathcal{L}_{joint} = \alpha \mathcal{L}_{1st} +  \mathcal{L}_{2nd} +  \nu  \mathcal{L}_{reg}
\end{equation}
where $\mathcal{L}_{1st}$ and $\mathcal{L}_{2nd}$ are loss functions to preserve first-order and second-order proximities respectively, while $\mathcal{L}_{reg}$ is the regularizer term. The authors propose a semi-supervised deep model to minimize the mentioned objective function. The deep model consists of two components: supervised and unsupervised. The supervised component attempts to preserve the first-order proximity while the unsupervised component attempts to preserve the second-order proximity by minimizing reconstruction loss of nodes. 

\textbf{Reproducibility notes}: We search for following  hyperparameters:  $\alpha$ = [1e-05, 0.2],  Penalty coefficient $\beta$ = [5, 10], Embedding dimension = [64, 128, 256].

\item \textbf{VAG \cite{kipf2016variational}}:
VAG is a graph autoencoder based network embedding method which minimizes the reconstruction loss of the adjacency matrix. The reconstruction matrix is generated as $\hat{A} = \sigma (ZZ^T)$ where $Z$ is node embeddings generated with Graph Convolutional Networks (GCN) \cite{kipf2017semi} as $Z = GCN(X,A)$ with $X$ as node features (see Table 1 for additional notation). In the case of unattributed graphs, the node feature matrix is the identity matrix. 

\textbf{Reproducibility notes}: We search for following  hyperparameters:  Epochs = [50, 100], Embedding dimension = [64, 128, 256].

\item \textbf{Watch Your Step \cite{abu2018watch}}: Watch Your Step (WYS) addresses the sensitivity issue of hyper-parameters in the random walk based embedding methods. WYS solves the sensitivity issue with the attention mechanism on the expected random walk matrix. The attention mechanism guides the random walk to focus on short or long term dependencies pertinent to the input graph. The objective function of WYS is given as
\vspace{-0.5em}
\begin{equation}
\begin{aligned}
\mathop{min}_{L,R,q} \quad & \beta \| q\|^2_2 -  \| E[D;q] \circ \log(\sigma(L*R^T)) \\ & -\mathbbm{1} [A = 0 ] \circ  \log(1- \sigma(L*R^T)) \|_1
\end{aligned}
\end{equation}
where $q$ is attention parameter vector, $L$ and $R$ are node embeddings, $E[D;q] $ is expectation on the random walk matrix (see Table \ref{notations} for rest of the notations).

\textbf{Reproducibility notes}: We search for following  hyperparameters: Learning rate =  [0.05, 0.1, 0.2, 0.5, 1.0], Number of Hops = 5, Embedding dimension = [64, 128, 256].

\end{enumerate}

%% file: datasets.tex
We select datasets from multiple domains, 
Table~\ref{tab:datasets} describes empirical properties of datasets. 
The selected datasets support both multi-label and multi-class classification. Directed as well as undirected datasets were selected in order to evaluate the embeddings methods on the link-prediction task efficiently.
Further, datasets with and without edge weights are also included, thereby, providing us with a comprehensive set of possibilities to evaluate the methods. 
We summarize the datasets below:
\begin{itemize}
\item Web: The WebKB datasets\footnote{http://linqs.cs.umd.edu/projects/projects/lbc/}~\cite{chakrabarti1998enhanced} consist of classified webpages (nodes) and hyperlinks between them (edges). Here, labels are the categories of the webpages.
\item Medical: The PPI dataset~\cite{breitkreutz2007biogrid} represents a subgraph of protein interactions in humans. Labels represent biological states corresponding to hallmark gene sets.
\item Natural Language: The Wikipedia dataset~\cite{mahoney2011large} is a dump of Wikipedia with nodes as words, edges corresponding to the co-occurrence matrix and labels corresponding to Part-of-Speech (POS) tags.
\item Social: The Blogcatalog dataset and Flickr dataset~\cite{zafarani2009social} represent social networks. Blogcatalog and Flickr both represent bloggers and their friendships. YouTube dataset \cite{grover2016node2vec} represents users and their friendships. Labels for Blogcatalog, Flickr, and YouTube correspond to the groups to which each user belongs. The Epinions  dataset \cite{richardson2003trust} represents user annotated trust relationships, where users annotate which other users they trust. These are used to determine the reviews shown to a user.

\item Citation: The DBLP, CoCit, and Pubmed datasets represent citation networks. DBLP (Co-Author) represents a subset of papers in DBLP\footnote{https://dblp.uni-trier.de/} from closely related fields. CoCit (Microsoft)~\cite{ms2016} corresponds to a co-citation subgraph of the Microsoft Academic Graph. Finally, Pubmed corresponds to a subset of diabetes-related publications on Pubmed\footnote{https://www.ncbi.nlm.nih.gov/pubmed/}. Labels in DBLP correspond to the sub-field of the paper. In CoCit, they correspond to the conference of the paper, and in Pubmed correspond to the types of diabetes.
\item Digital: The p2p-Gnutella dataset~\cite{ripeanu2002mapping} represents connections between hosts on a peer-to-peer file sharing network. This dataset has no node labels.
\item Voting: The Wiki-Vote dataset~\cite{leskovec2010signed} is constructed from voting data in multiple elections for Wikipedia administratorship. Users are nodes, and (directed) edge ($i$, $j$) represents a vote from user $i$ to user $j$. This dataset also has no node labels.
\end{itemize}

\begin{table}[]
    \centering
    \resizebox{1.0\linewidth}{!}{
    \begin{tabular}{lrrrccc}
    \toprule
    \textbf{Dataset}            & \textbf{\#Nodes} & \textbf{\#Edges} & \textbf{\#Labels} &  \textbf{(C/L)\tnote{1}} & \textbf{D} & \textbf{W} \\
    \midrule
    WebKB (Texas)      & 186             & 464            & 4                 & C                                & F        & T     
    \\
    WebKB (Cornell)    & 195             & 478             & 5                 & C                                & F        & T     
    \\
    WebKB (Washington) & 230             & 596             & 5                 & C                                & F        & T     
    \\
    WebKB (Wisconsin)  & 265             & 724            & 5                 & C                                & F        & T     
    \\
    PPI                & 3,890            & 38,739           & 50                & L                                & F        & F     
    \\
    Wikipedia          & 4,777            & 92,517          & 40                & L                                & F        & T    
    \\
    Blogcatalog        & 10,312           & 333,983          & 39                & L                                & F        & F     
    \\
    DBLP (Co-Author)   & 18,721           & 122,245          & 3                 & C                                & F        & T     
    \\
    CoCit (Microsoft)  & 44,034           & 195,361          & 15                & C                                & F        & F     
    \\
    Wiki-Vote         & 7,115            & 103,689          & -               & -                                & T        & F       
    \\
    Pubmed             & 19,717           & 44,338           & 3                 & C                                & T        & F    
    \\
    p2p-Gnutella      & 62,586           & 147,892          & -                 & -                                & T        & F     
    \\
    Flickr             & 80,513           & 5,899,882        & 195               &  L                               & F        & F     
    \\
    Epinions            & 75,879           & 508,837            & - & -   & T & F \\
    YouTube            & 1,134,890          & 2,987,624            & 47 & C   & F & F \\ 
    \bottomrule 
    \end{tabular}
    }
    \caption{Dataset Properties. ``C"/``L" refers to Multi Class or vs Multi Label(``L"). ``D" refers to Directed and ``W" refers to weighted.}\label{tab:datasets}
    \vspace{-1em}
\end{table}


%% file: experimental_setup.tex
In this section, we elaborate on the experimental setup for link prediction and node classification tasks employed to evaluate the quality of embeddings generated by different methods. We present two heuristics baselines for both the tasks and define the metrics used for comparing the embedding methods.


\subsection{Link Prediction}
\label{lp_setup}

Prediction of ties is an essential task in multiple domains where the relational information is costlier to obtain such as drug-target interactions \cite{Crichton2018},  protein-protein interactions \cite{Airoldi06mixedmembership}, or when the environment is partially observable. 
 The problem of prediction of tie/link between two nodes $i$ and $j$ is often evaluated in one of two ways. The first is to treat the problem as a binary classification problem. The second is to use the dot product on the embedding space as a scoring function to evaluate the strength of the tie. 

The edge features for binary classification consists of node embeddings of nodes $i$ and $j$, where two node embeddings are aggregated with a binary function. In our study, we experimented with three binary functions on node embeddings: Concatenation, Hadamard, and L2 distance. 
 We used logistic regression as our base classifier for the prediction of the link. The parameters of the logistic regression are tuned using GridSearchCV with 5-fold cross validation with scoring metric as `roc\_auc'.  We evaluate the link prediction performance with metrics: Area Under the Receiver Operating Characteristics (AUROC) and Area Under Precision-Recall curve (AUPR). An alternative evaluation strategy is to predict the presence of link $(i, j)$ based on dot product value of node embeddings of nodes $i$ and $j$. We study the impact of both the evaluation strategies in Section \ref{lp_results}.

\textbf{Construction of the train and test sets}: The method of construction of train and test sets for link prediction task is crucial for comparison of embedding methods. 
The train and test split consists of 80\% and 20\% of the edges respectively and are constructed in the following order: 

\vspace{-0.5em}
\begin{enumerate}[noitemsep]
    \item Self-loops are removed.
    \item We randomly select 20\% of all edges as positive test edges and add them in the test set.
    \item Positive test edges are removed from the graph. We find the largest weakly connected component formed with the non-removed edges. The edges of the connected component form positive train edges.
    \item We sample negative edges from the largest weakly connected component and add the sampled negative edges to both the training set and test set. The number of negative edges is equal to the number of positive edges in both training and test sets.
    \item For directed graphs, we form ``directed negative test edges" which satisfy the following constraint: $(j, i) \not\in E$ but $(i, j) \in E$ where $E$ refers to edges in the largest weakly connected component. We add the directed negative test edges $(j, i)$ edges to our test set. The number of ``directed negative test edges" is around 10\% of negative test edges in the test set.
    \item Nodes present in the test set, but not present in the training set, are deleted from the test set.
    
\end{enumerate}

In case of large datasets ($>$5M edges), we reduce our training set. We consider 10\% of both randomly selected positive and negative train edges for learning the binary classifier. The learned model is evaluated on the test set. The above steps are repeated for 5 folds of a train:test splits of 80:20\% and we report the average AUROC and AUPR scores across 5 folds. 

\subsection{Node classification}
In network embedding literature, node classification is the most popular way of comparing the quality of embeddings generated by different embedding methods. The generated node embeddings are treated as node features, and node labels are treated as ground truth. The classification task performed in our experiments is either Multi-label or Multi-class classification. The details on the classification task performed on each dataset are provided in Table \ref{tab:datasets}. We select Logistic Regression as our classifier. The hyperparameters of the logistic regression are tuned using GridSearchCV with 5-fold cross validation with scoring metric as `f1\_micro'. 
We split the dataset with  50:50 train-test splits. 
The learned model is evaluated on the test set, and we report the results averaged over 10 shuffles of train-test sets. The model does not have access to test instances while training. 

We note that a majority of the efforts in the literature do not tune the hyper-parameters of Logistic Regression. Default hyper-parameters are not always the best hyper-parameters for Logistic Regression. For instance, with default hyper-parameters of LR classifier, the Macro-f1 performance of Laplacian eigenmaps on Blogcatalog dataset is 3.9\% for the train-test split of 50:50. 
However, tuning the hyper-parameters results in significant improvement of Macro-f1 score to 29.2\%. 

The choice of a ``linear" classifier to evaluate the quality of embeddings is not a hard constraint in the node classification task. In this work, we also test the idea of leveraging a ``non-linear" classifier for the node classification task and use EigenPro \cite{ma2017diving} classifier for the same task. On large datasets, EigenPro provides a significant performance boost over the state-of-the-art kernel methods with faster convergence rates \cite{ma2017diving}. In the experiments, we see a benefit to this approach, up to 15\% improvement in Micro-f1 scores with non-linear classifier compared to the linear classifier.



\subsection{Heuristics}
Next, we present heuristics baseline for both link prediction and node classification tasks. The purpose of defining heuristics baseline is to assess the difficulty of performing a particular task on a particular dataset and also to compare the performance of sophisticated network embedding methods over simple heuristics.

\subsubsection {Link Prediction Heuristics}
\label{lp_heuristics}

In the link prediction literature, there exist multiple similarity based metrics \cite{lu2011link} which can predict a score for link formation between two nodes. Examples of such metrics include Jaccard Index \cite{jaccard1901etude,wangsatuluri07}, Adamic Adar \cite{adamic2003friends}. These similarity-based metrics often base their predictions on the neighborhood overlap between the nodes. We combine the similarity-based metrics to form a curated feature vector of an edge \cite{10.1007/978-3-030-05411-3_7}. The binary classifier in the link prediction task is then trained on the generated edge embeddings. Our selected similarity-based metrics are Common Neighbors (CN), Adamic Adar (AA) \cite{adamic2003friends}, Jaccard Index (JA) \cite{jaccard1901etude}, Resource Allocation Index (RA) \cite{zhou2009predicting} and Preferential Attachment Index (PA) \cite{barabasi1999emergence}. The similarity-based metrics CN, JA, and PA captures first-order proximity between nodes, while the metrics AA and RA capture second-order proximity between nodes. We found this heuristic based model to be highly competitive as compared to the embedding methods on multiple datasets.

\subsubsection {Node Classification Heuristics}

Nodes in the graph can be characterized/represented by their properties.  We combine the node properties to form a feature vector/embedding of a node.  The classifier in node classification task is then trained on the generated node embeddings.
The node properties capture information such as nodes' neighborhood, influence on other nodes, structural properties. We select following node properties :  Degree, PageRank \cite{page1999pagerank}, Clustering Coefficient, Hub and Authority scores \cite{kleinberg1999authoritative}, 
Average Neighbor Degree, and Eccentricity \cite{newman2010networks}.
We treat the graph as undirected while computing the node properties. As the magnitude of each node property varies with another, we perform column-wise normalization with RobustScaler available from Scikit-learn. We will show in the experiments Section  \ref{nc_results} that the node classification heuristics baseline is competitive with most of the embedding methods on datasets with fewer labels.


\subsection{Comparison Measures}
In this section, we present two measures for comparing the performance of embedding methods in the downstream machine learning task. 

\textbf{\textit{Mean Rank}}: We compute the rank of all the embedding methods on each dataset based on selected performance metric and report the average rank of an embedding method across all datasets as the \textit{Mean Rank} of the embedding method. Let $R_{e,d}$ be the rank of embedding method $e$ on dataset $d$ with $D$ being the set of datasets then  mean rank of embedding method $MR_e$ is given by

\begin{equation}
    MR_e = \frac{\sum_{d \in D} R_{e,d}}{ |D| }
\end{equation}

\textbf{\textit{Mean Penalty \cite{vijayan2018fusion}}}: We define \textit{penalty} of an embedding method $e$  on a dataset $d$ as difference between best score  achieved by any embedding method on dataset $d$ and the score achieved by embedding method $e$ on same dataset $d$. Score is the selected performance metric for a particular downstream ML task. Let $E$ be the set of embedding methods and $S_{e,d}$ be the score achieved by embedding method $e$ on same dataset $d$, then the \textit{Mean Penalty} $MP_e$ is given by

\begin{equation}
    MP_e  = \frac{\sum_{d \in D} max(\{S_{e',d}\}) - S_{e,d}}{ |D| } \;\;\;\; ; e' \in E
\end{equation}

For a model, lower values for \textit{Mean Rank} and \textit{Mean Penalty} suggest better performance.  We compare the embedding methods with \textit{Mean Rank}, and \textit{Mean Penalty} measures on the datasets where all the embedding methods complete execution. Though the measures do not consider the dataset size or missing values, the measures are simple and intuitive. 


%% file: experimental_results.tex
In this section, we report the performance of network embedding methods on link prediction task and node classification task. We tune both the parameters of embedding methods and the parameters of classifiers in link prediction and node classification task. Whenever possible, we rely on the authors' code implementation of the embedding method.  All the methods which do not complete execution on large datasets are executed on a modern machine with  500  GB  RAM  and 28 cores. All the evaluation scripts are executed in the same virtual python environment.\footnote{The evaluation scripts and datasets are available at https://github.com/PriyeshV/NRL\_Benchmark.}

\input{LP_results.tex}

\input{NC_results.tex}


%% file: LP_results.tex
\begin{figure*}[htb]
\centering
\begin{subfigure}{.55\textwidth}
  \centering
  \includegraphics[width=1.0\linewidth, height=14em]{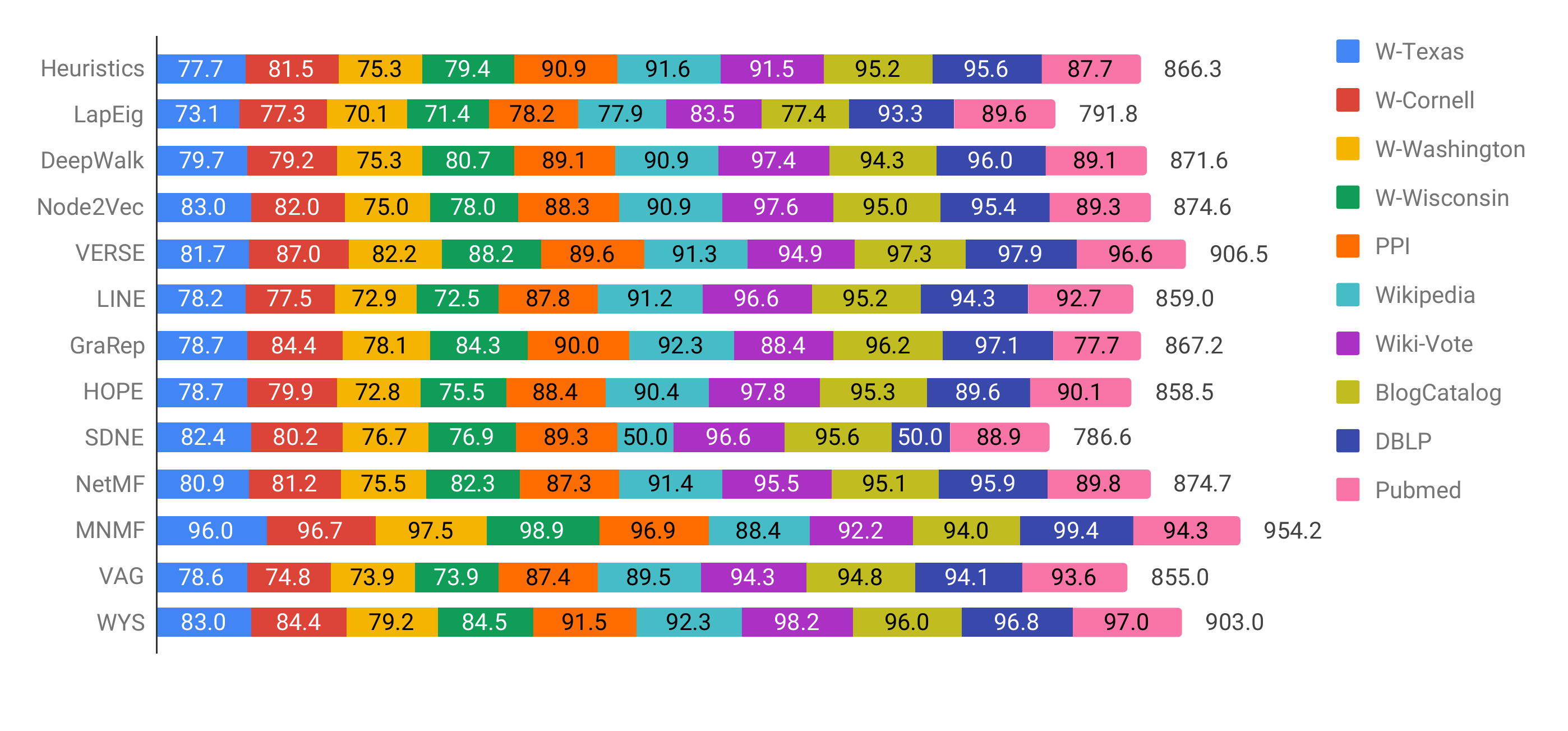}
  \caption{Smaller datasets: All methods complete execution.}
  \label{fig:lp_auc_scalable}
\end{subfigure}%
\begin{subfigure}{.45\textwidth}
  \centering
  \includegraphics[width=1.0\linewidth, height=14em]{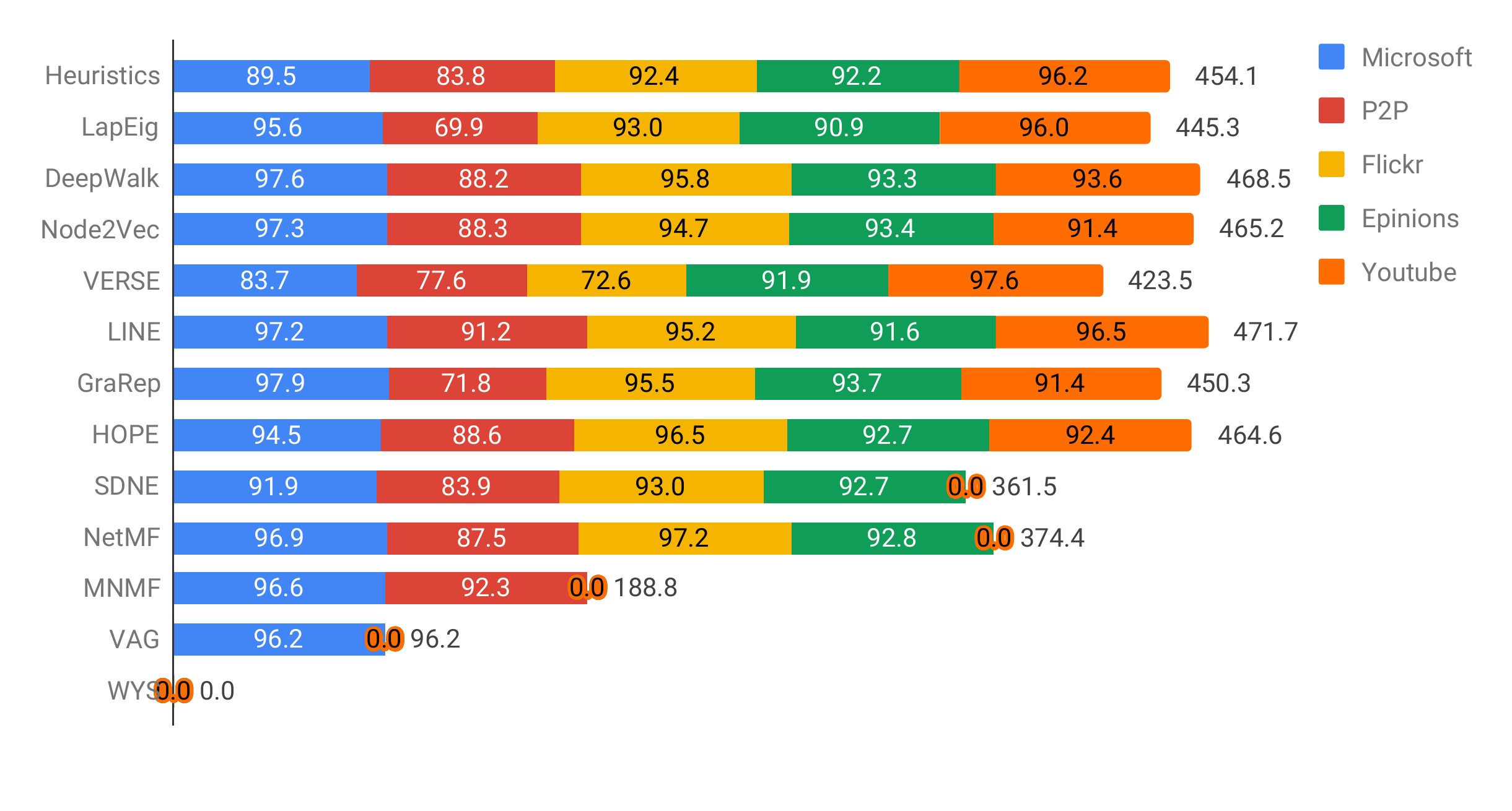}
  \caption{Larger datasets: Not all methods complete execution.}
  \label{fig:lp_auc_large}
\end{subfigure}
\caption{Link Prediction performance measured with \textbf{AUROC}. Figure \ref{fig:lp_auc_scalable} shows the best average AUROC of all methods on datasets where all methods' results are available. Not all methods are scalable on the datasets shown in Figure \ref{fig:lp_auc_large}. If a method runs out of memory or faults on a particular dataset, we represent its performance as 0.0 in the above plot.}
   \label{fig:LP-AUC} 
\end{figure*}

\begin{figure*}[ht]
\centering
\begin{subfigure}{.55\textwidth}
  \centering
  \includegraphics[width=1.0\linewidth, height=14em]{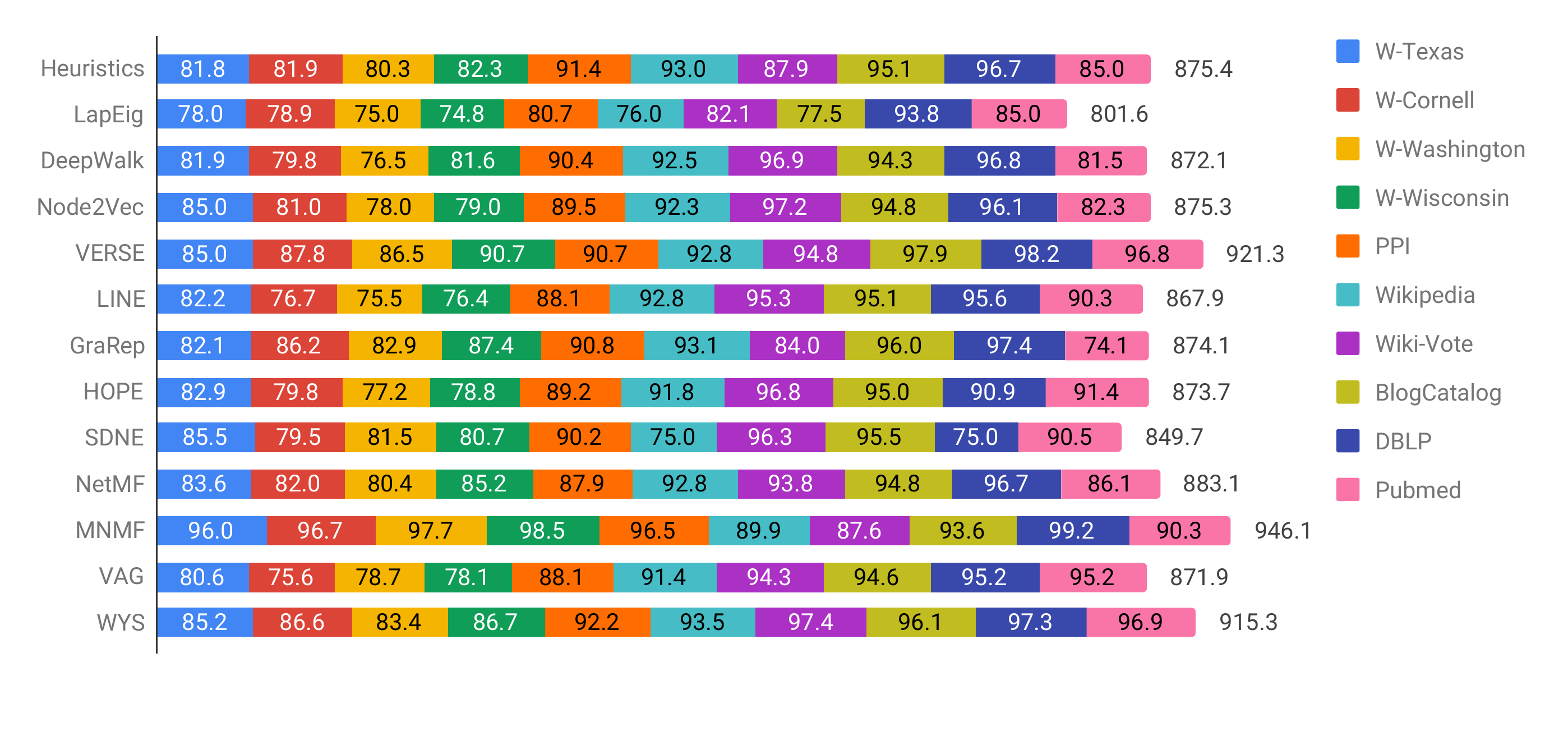}
  \caption{Smaller datasets: All methods complete execution.}
  \label{fig:lp_aupr_scalable}
\end{subfigure}%
\begin{subfigure}{.45\textwidth}
  \centering
  \includegraphics[width=1.0\linewidth, height=14em]{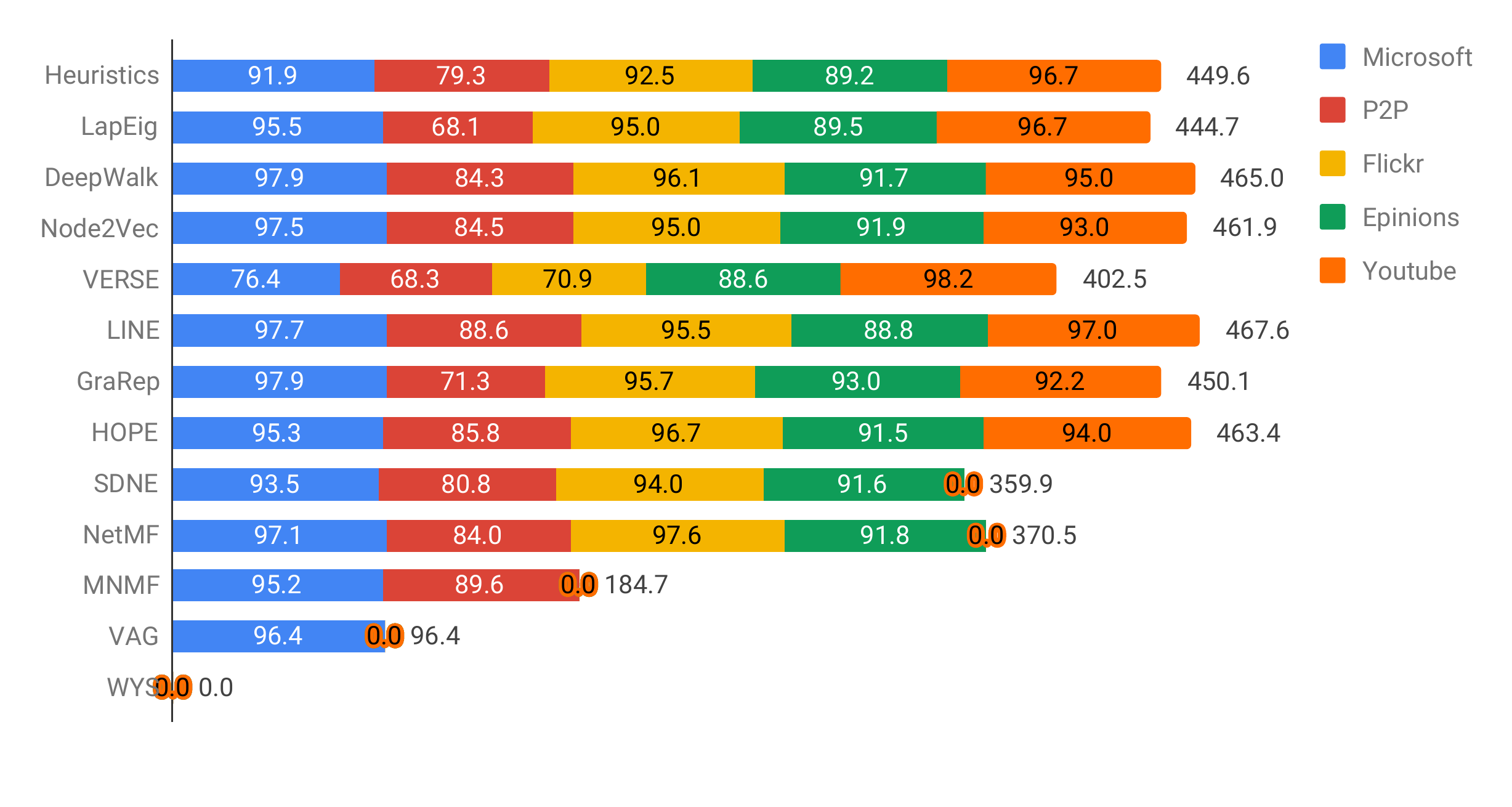}
  \caption{Larger datasets: Not all methods complete execution.}
  \label{fig:lp_aupr_large}
\end{subfigure}
\caption{Link Prediction performance measured with \textbf{AUPR}. Figure \ref{fig:lp_auc_scalable} shows the best average AUPR of all methods on datasets where all methods' results are available. Not all methods are complete execution on the datasets shown in Figure \ref{fig:lp_auc_large}. If a method runs out of memory or faults on a particular dataset, we represent its performance as 0.0 in the above plot.}
   \label{fig:LP-AUPR} 
\end{figure*}

\begin{figure}[!ht]
    \centering
    \includegraphics[width=1.0\linewidth]{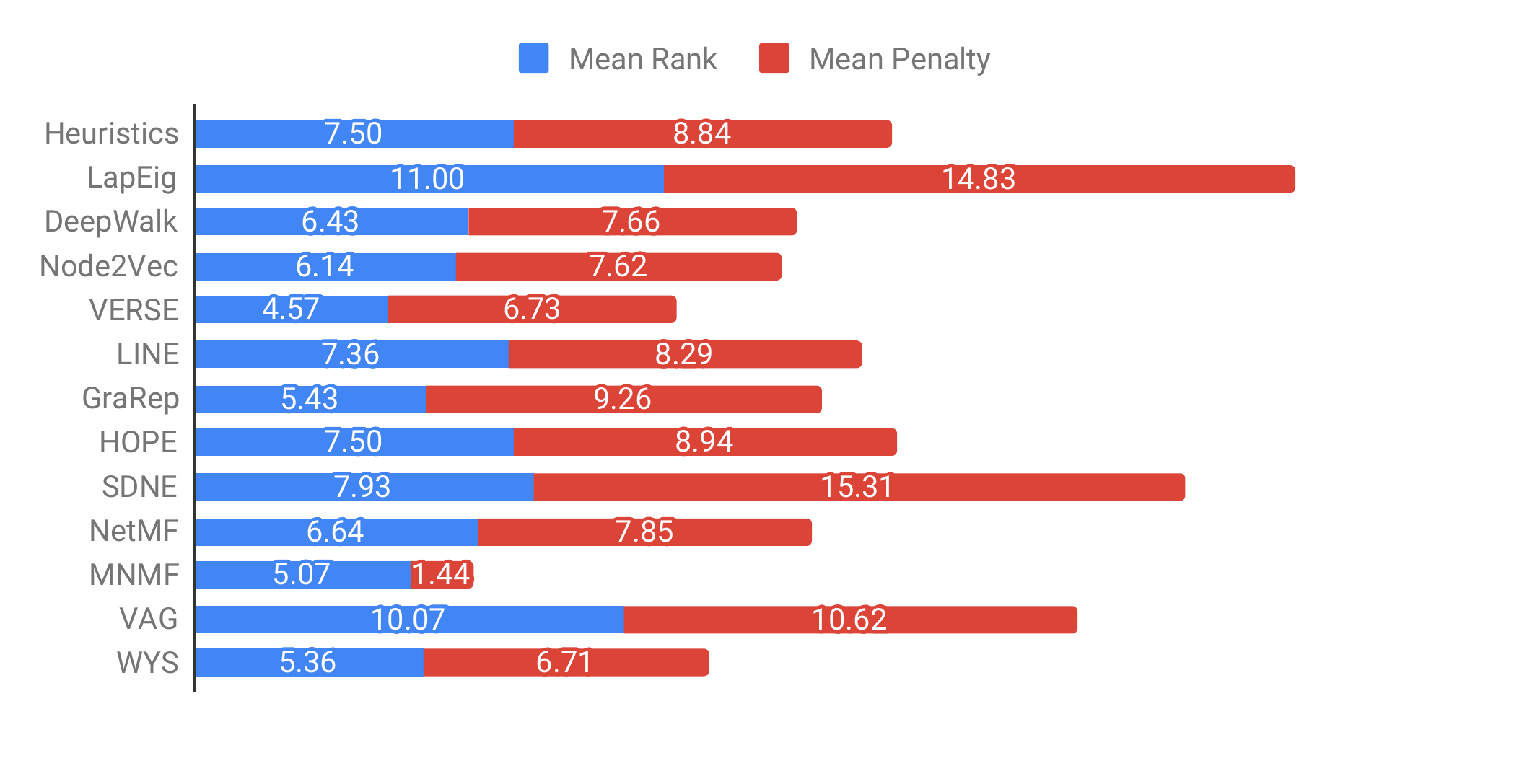}
    \caption{Mean Rank and Mean Penalty for link prediction with selected performance metric as AUROC.}
    \label{fig:mr_op_lp}
\end{figure}

\begin{figure}[t]
\centering
\includegraphics[width=1.0\linewidth]{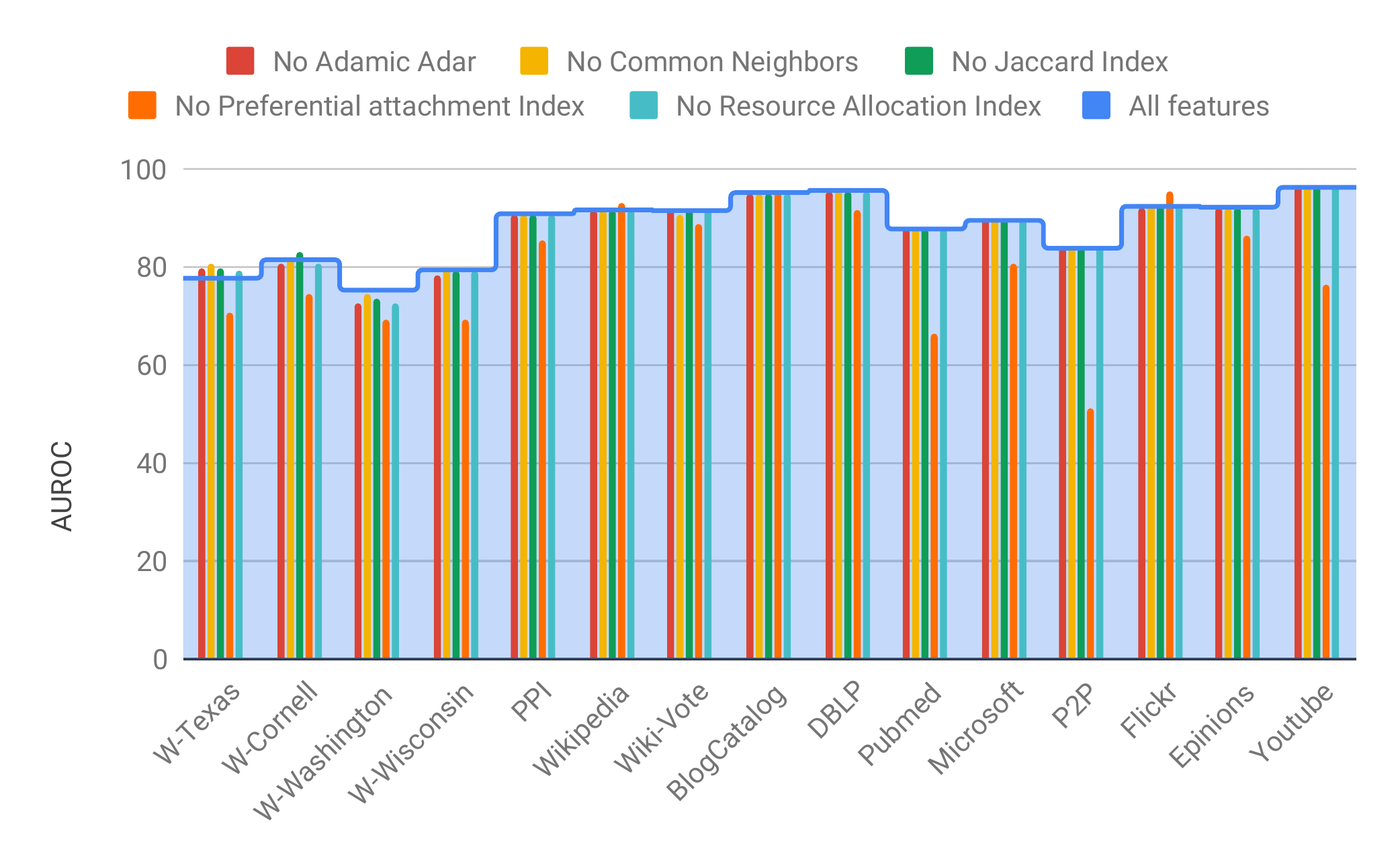}
\caption{Feature study on the link prediction for link prediction heuristics.}
\label{fig:lp_heu_ablation}
\end{figure}

\begin{figure}[t]
    \centering
     \includegraphics[width=\linewidth]{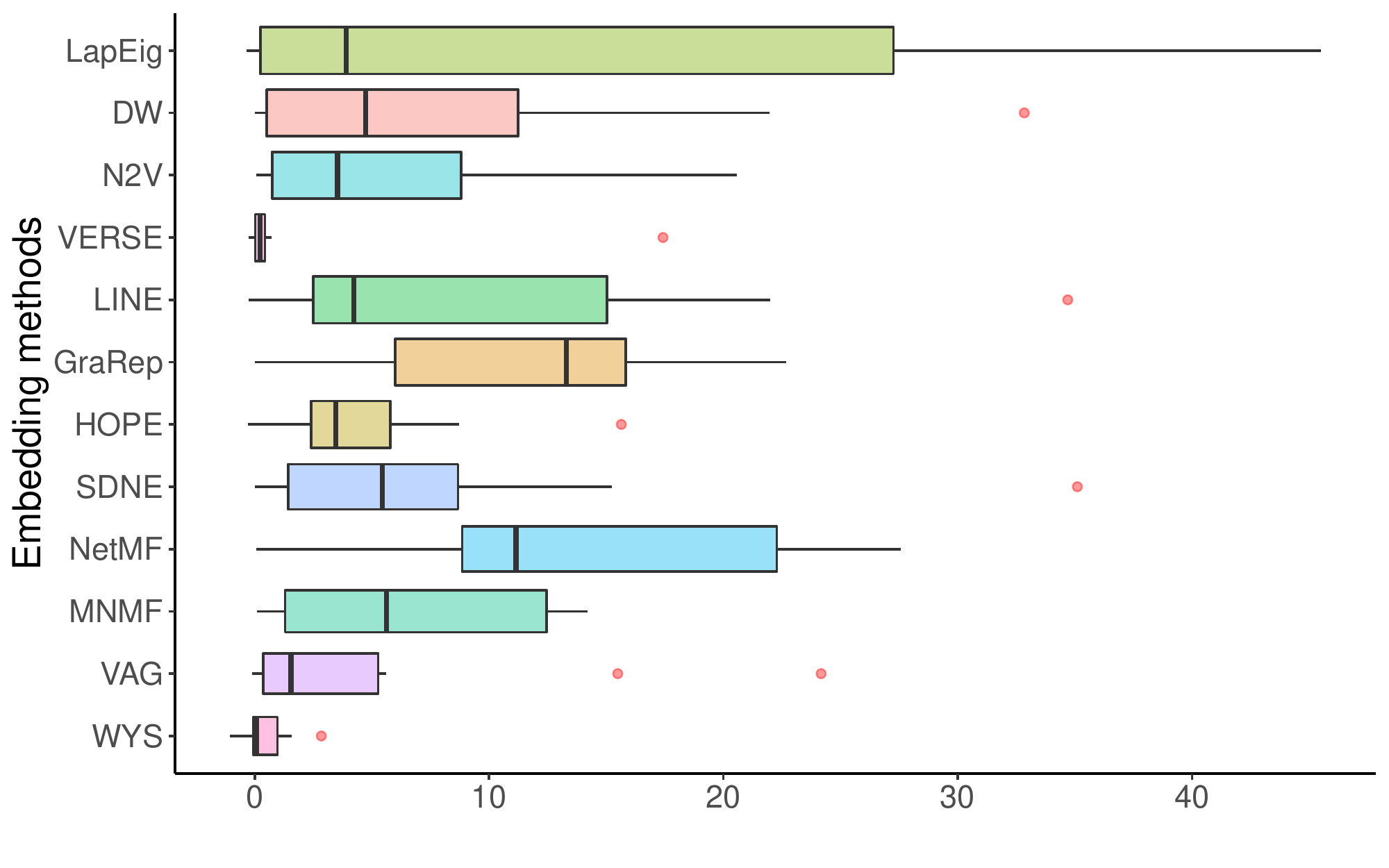}
    \caption{The box-plot represents distribution of the differences between computed AUROC score with classifier and computed AUROC score with dot product on all the datasets. The difference is statistically significant (paired t-test) for all methods, except for Verse and WYS, with significance level 0.05.}
    \label{fig:classi_vs_dot_product}
\end{figure}

\begin{figure}[!ht]
    \centering
    \includegraphics[scale=0.35]{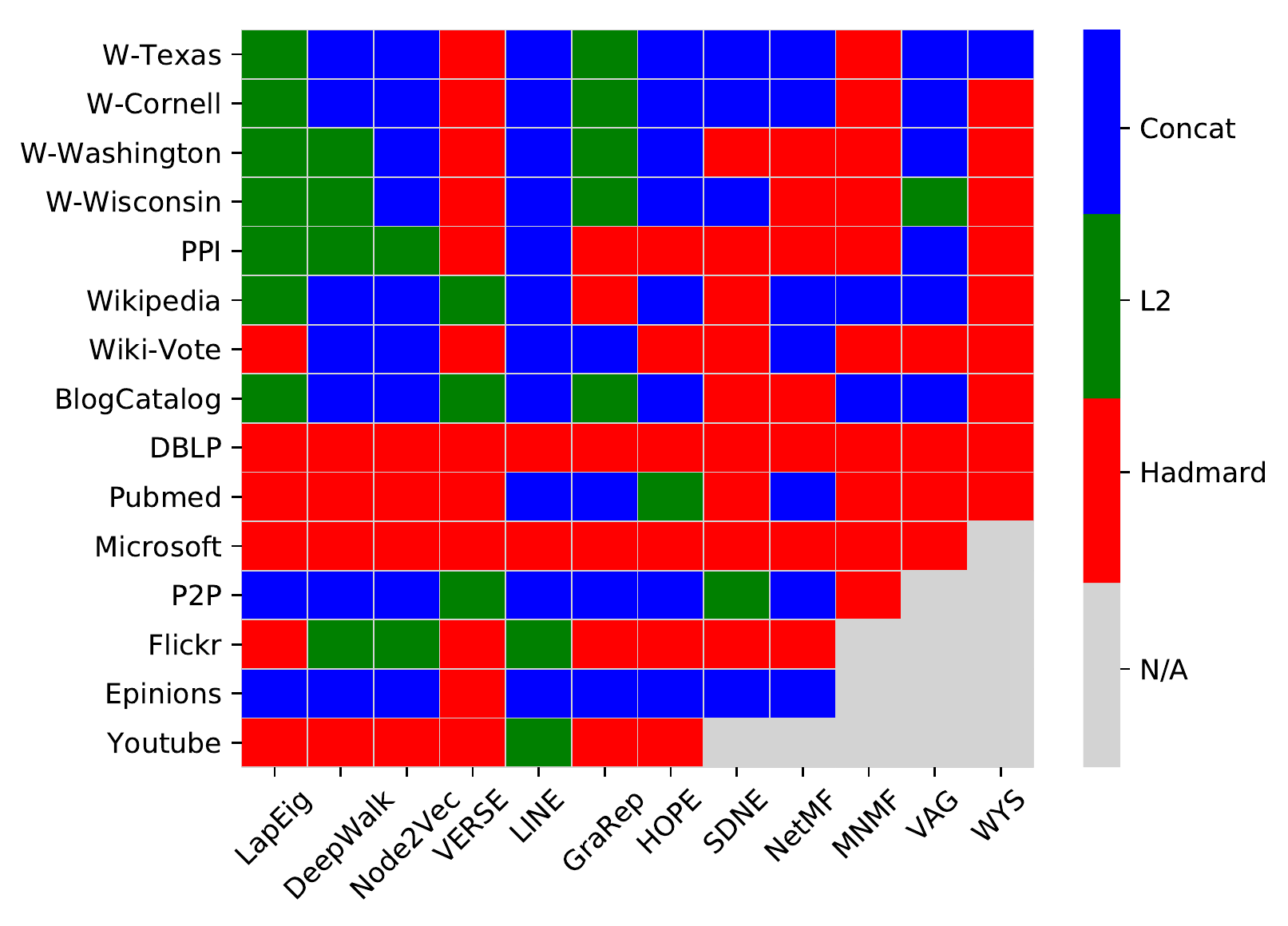} 
    \caption{The heatmap shows which binary function -- Hadamard, Concatenation and L2 -- resulted in best average AUROC score for an embedding method on a particular dataset.}
    \label{fig:lp_features}
\end{figure}

\begin{figure}[t]
    \centering
     \includegraphics[width=\linewidth]{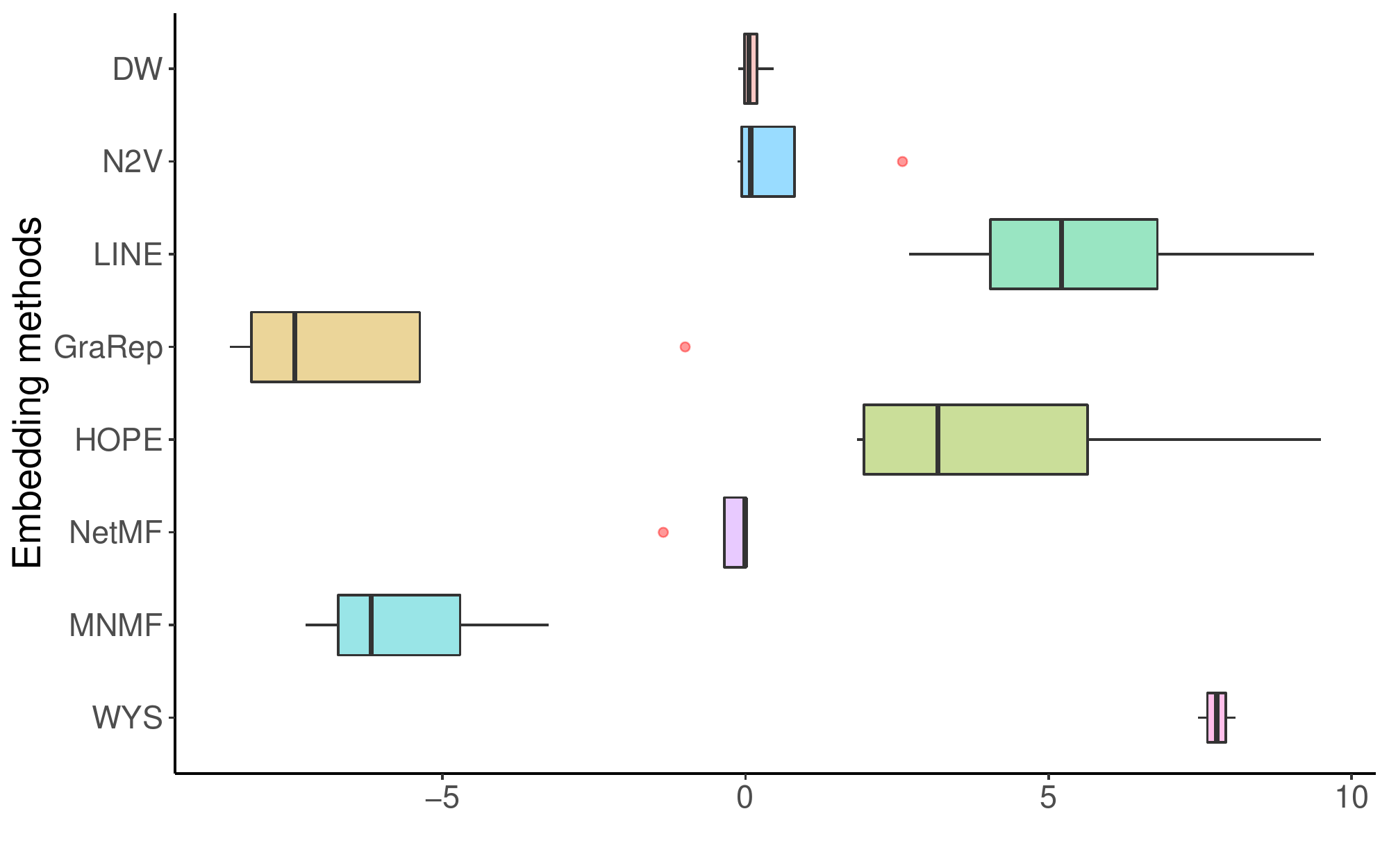}
    \caption{The box-plot represents the distribution of the differences between computed AUROC score with node + context embeddings and computed AUROC score with only node embeddings on directed datasets. The difference is statistically significant (paired t-test) for methods LINE, WYS, GraRep and MNMF with significance level 0.05.}
    \label{fig:cxt_no_cxt}
\end{figure}

\begin{figure*}[t]
\centering
\begin{subfigure}{.5\textwidth}
\includegraphics[width=1.0\linewidth]{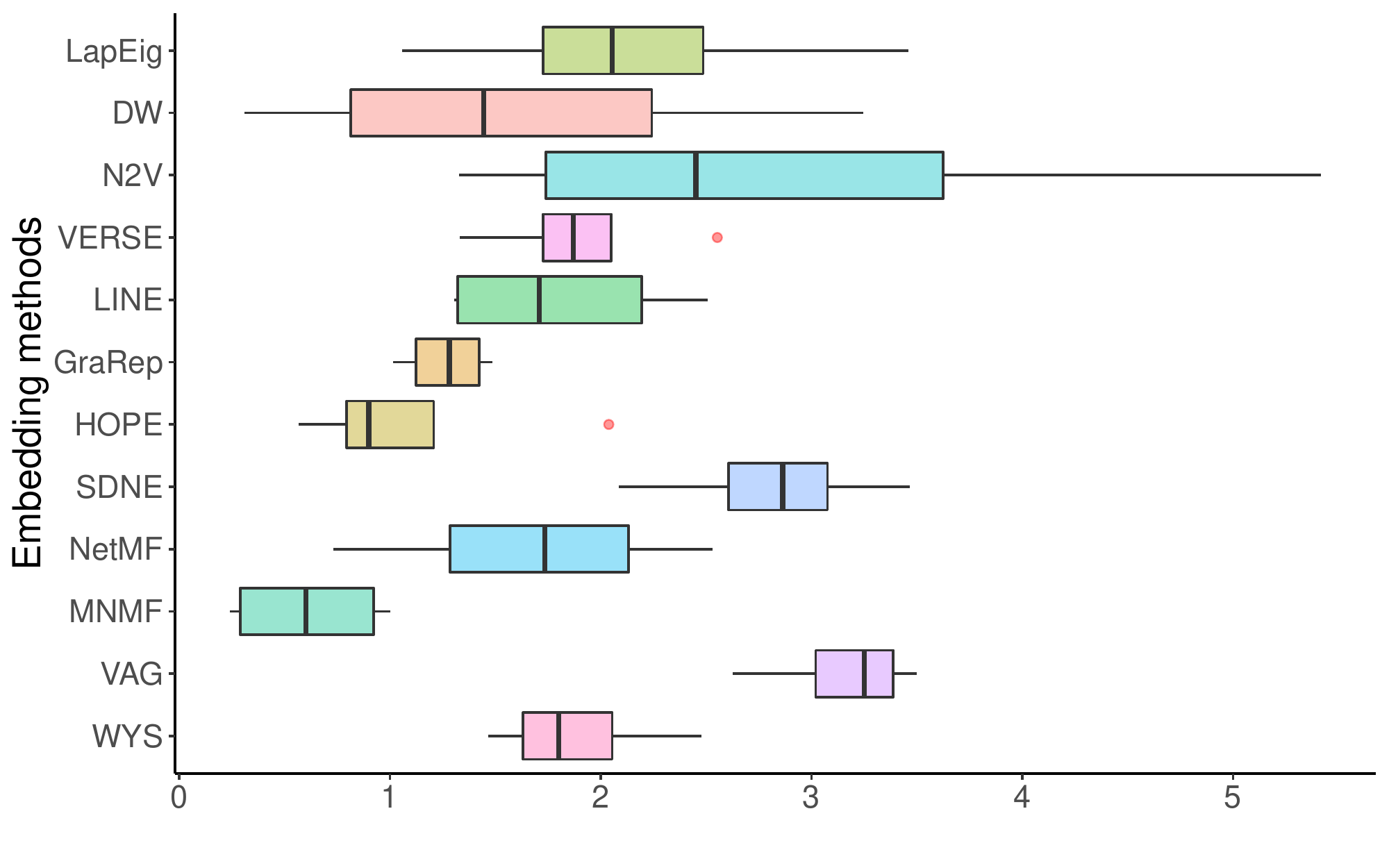}
  \caption{WebKB datasets}
  \label{fig:lp_std_webkbonly}
\end{subfigure}%
\begin{subfigure}{.5\textwidth}
\includegraphics[width=1.0\linewidth]{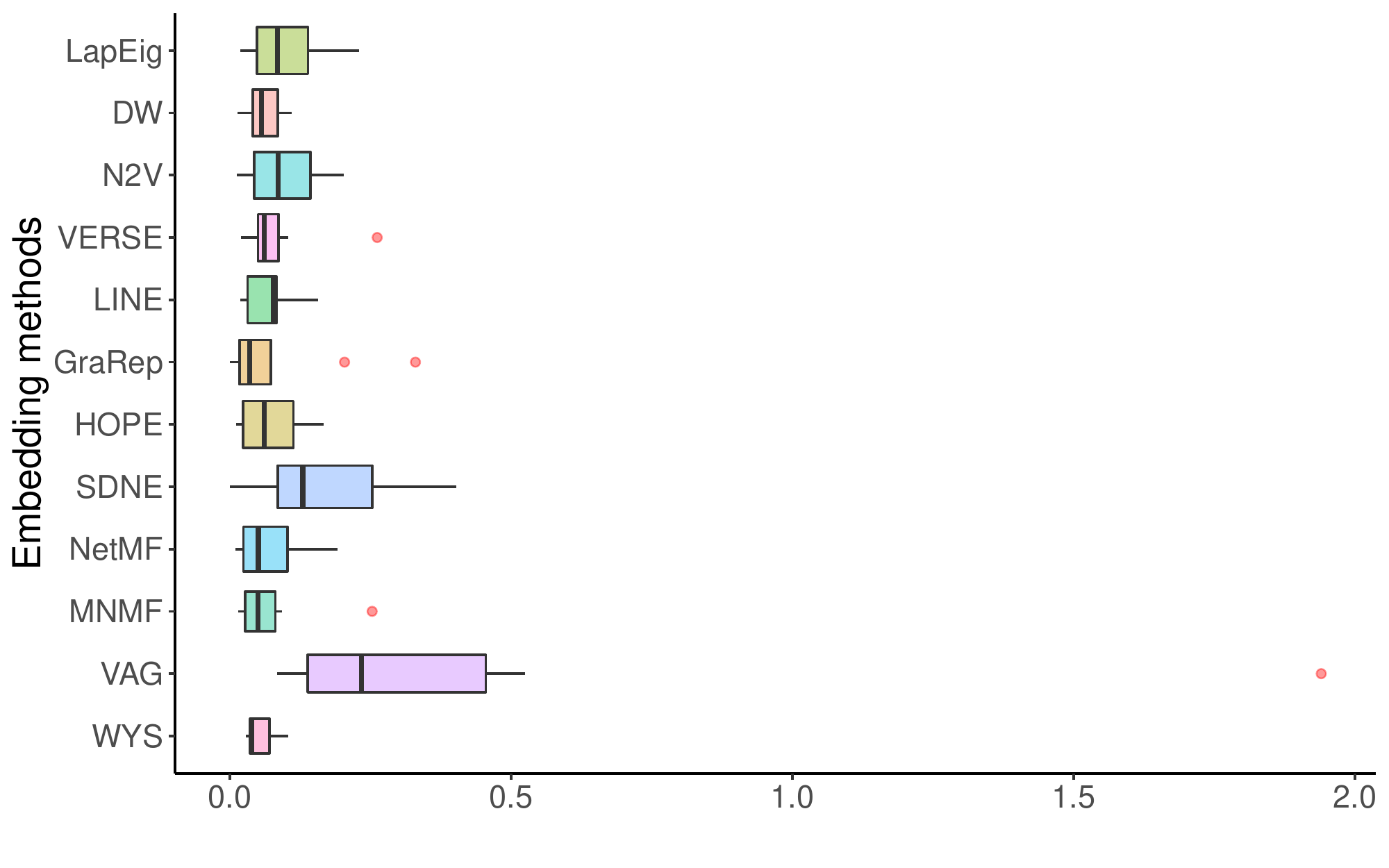}
  \caption{All datasets except WebKB datasets}
  \label{fig:lp_std_webkbno}
\end{subfigure}
\caption{The box-plot represents the distribution of the standard error of AUROC scores computed on all the datasets for the link prediction task.}
\label{fig:stddev}
\end{figure*}

\begin{figure}[t]
\centering
\includegraphics[width=1.0\linewidth]{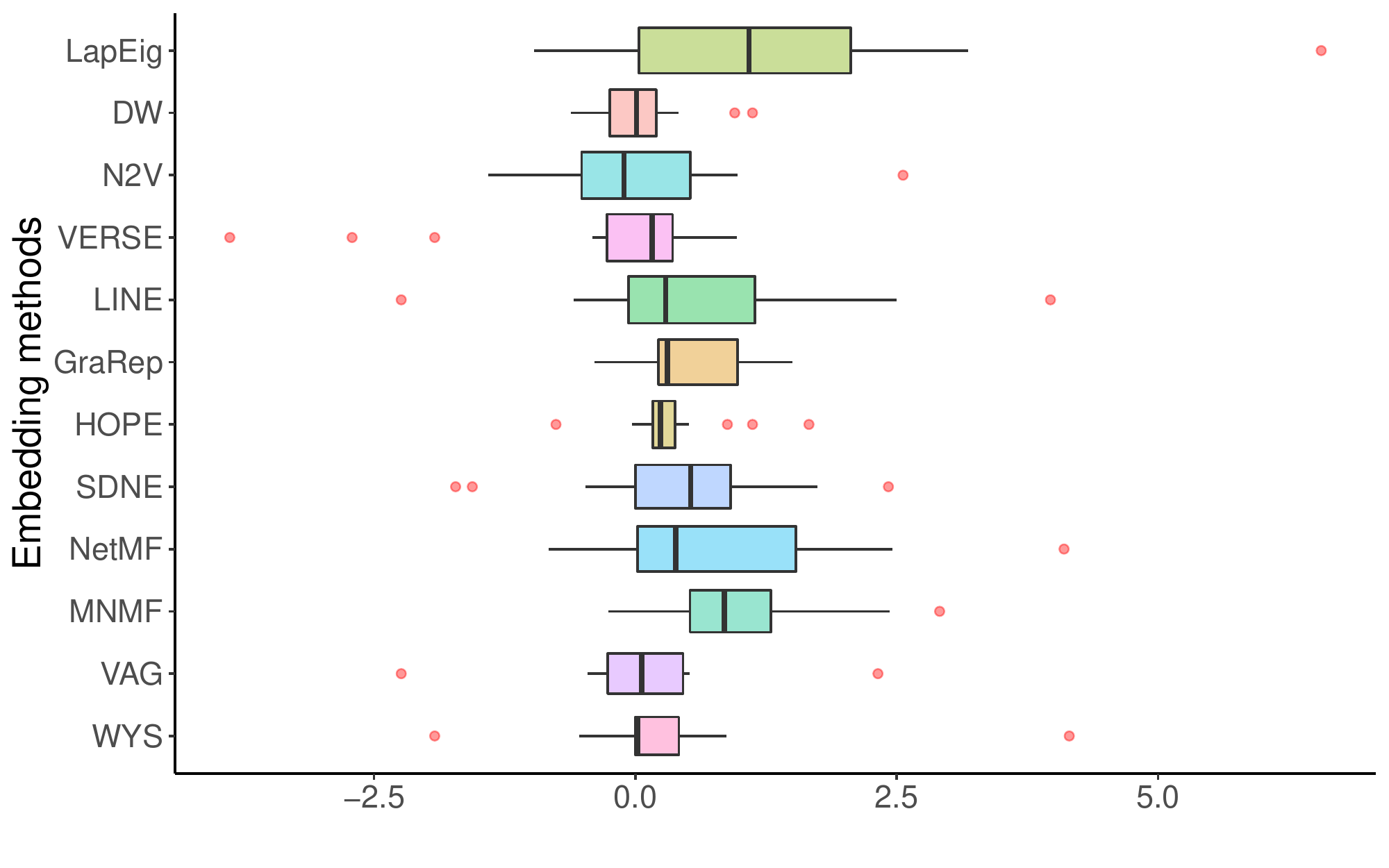}
\caption{The box-plot represents the distribution of the differences in AUROC scores between 128 dimensional embedding and 64 dimensional embedding for each method on all datasets}
\label{fig:lp_dim}
\end{figure}

\begin{figure*}[t]
\centering
\begin{subfigure}{.48\textwidth}
\includegraphics[width=1.0\linewidth]{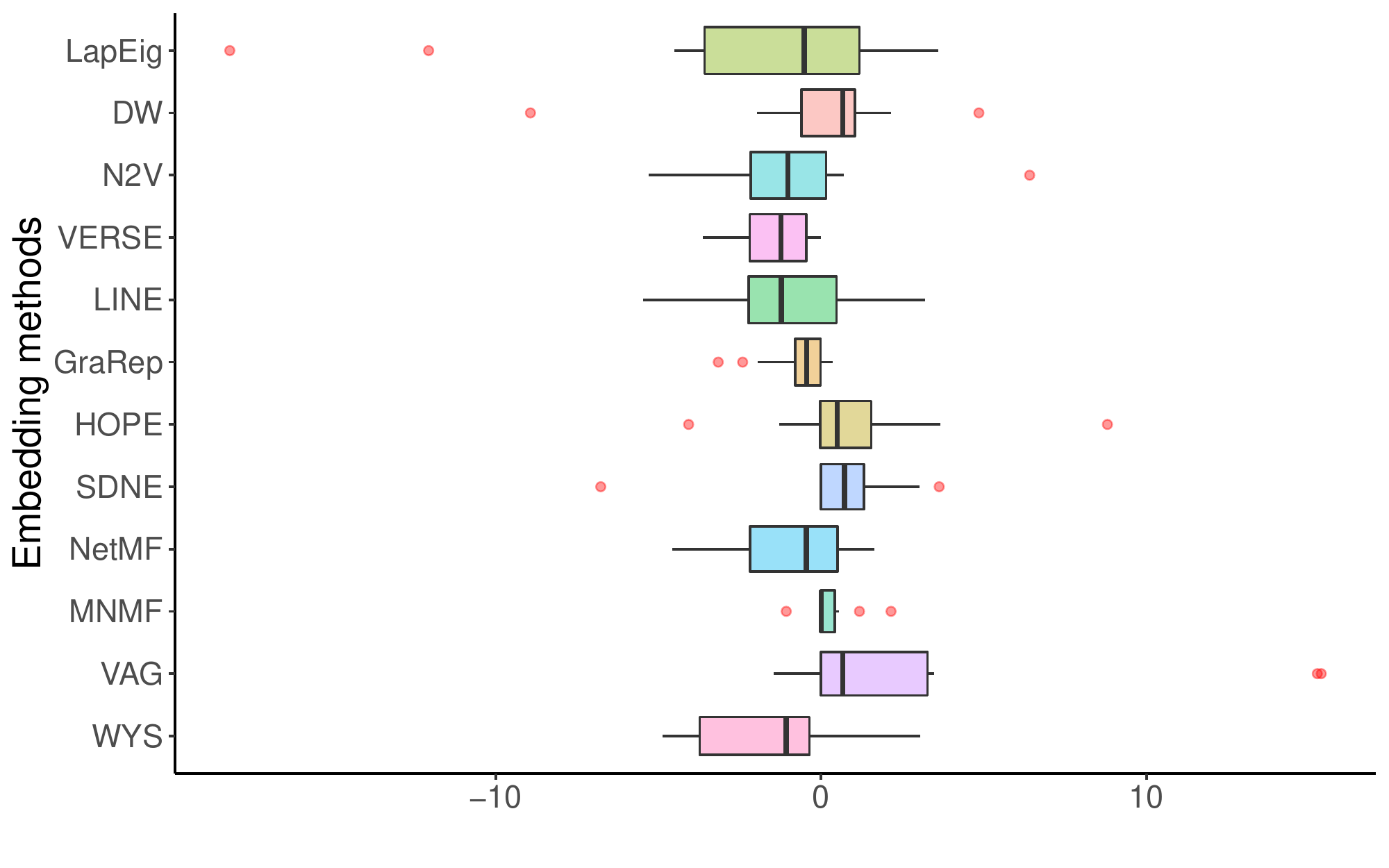}
\caption{Prediction of link through classifier}
\label{fig:lp_l2norm}
\end{subfigure}%
\begin{subfigure}{.48\textwidth}
\includegraphics[width=1.0\linewidth]{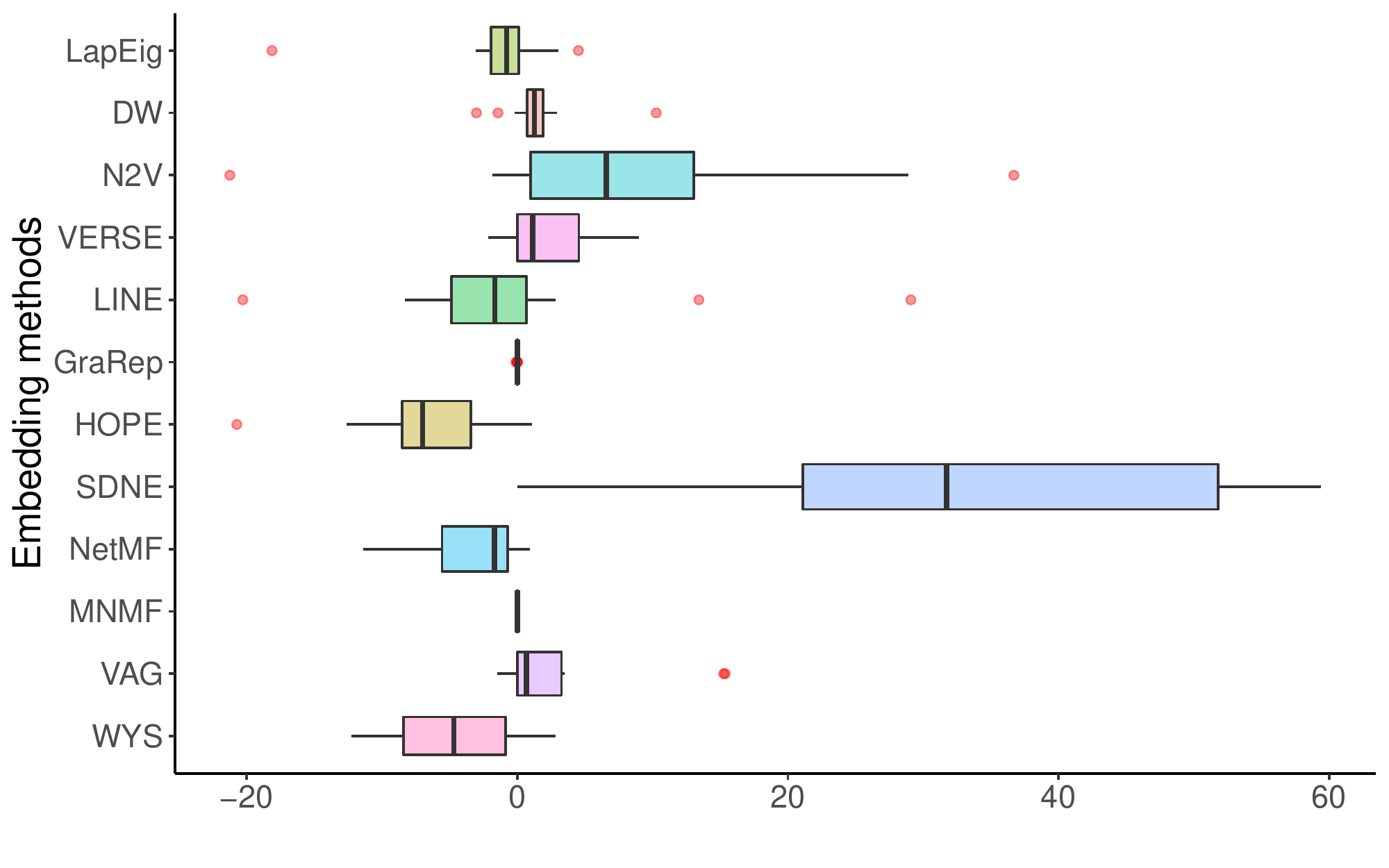}
\caption{Prediction of link through dot-product}
\label{fig:auc_lp_l2norm_lr}
\end{subfigure}
 \caption{The box-plot represents the distribution of the differences in AUROC scores between L2 normalized embeddings and unnormalized embeddings}
 \label{fig:lp_norm}
\end{figure*}

\begin{figure*}[t]
\centering
\begin{subfigure}{.5\textwidth}
  \centering
  \includegraphics[width=1.0\linewidth]{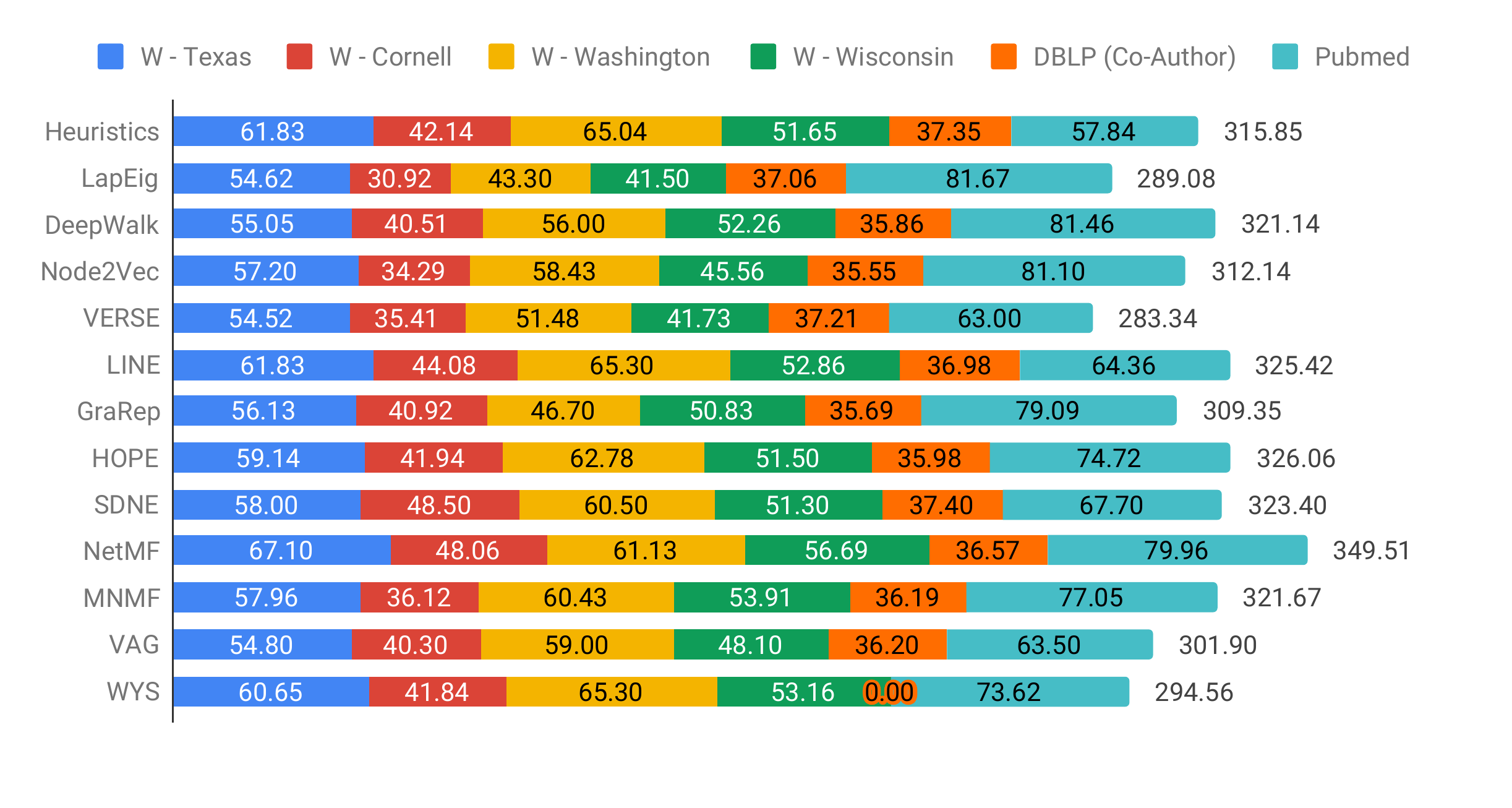}
  
  \caption{Datasets with few labels}
  \label{fig:micro_few_label}
  
\end{subfigure}%
\begin{subfigure}{.5\textwidth}
  \centering
\includegraphics[width=1.0\linewidth]{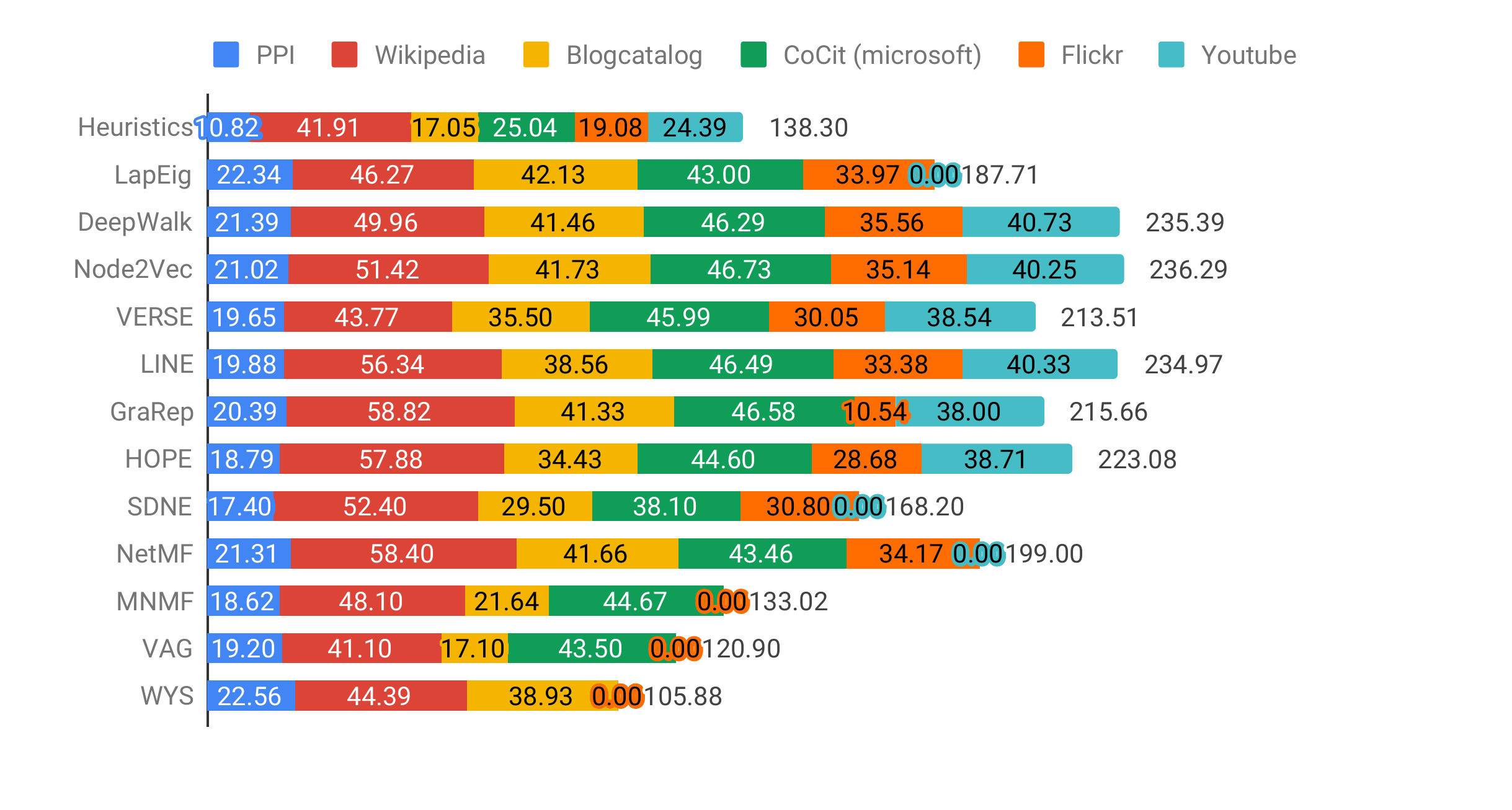}
  \caption{Datasets with more labels}
  \label{fig:micro_more_label}
\end{subfigure}
\caption{The node classification performance measured  with Micro-f1 on train-test split of 50:50  with Logistic Regression. For each method, the number at the end of bar represent the summation of the Micro-f1 values across the datasets.  }
\label{fig:node_micro}
\vspace{-1em}
\end{figure*}

\begin{figure*}[t]
\centering
\begin{subfigure}{.5\textwidth}
  \centering
\includegraphics[width=1.0\linewidth]{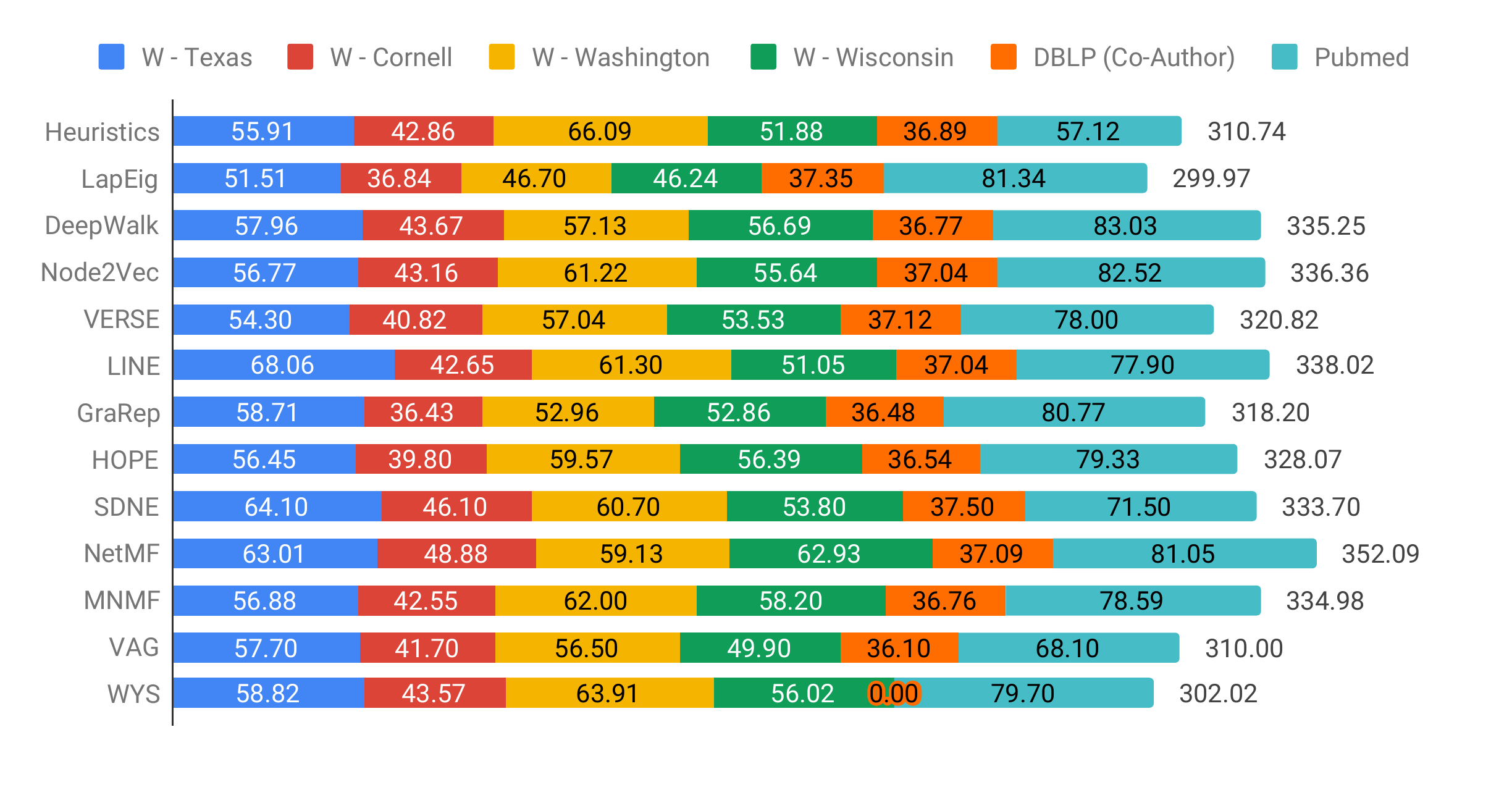}
  \caption{Datasets with few labels}
  \label{fig:micro_few_label_nl}
\end{subfigure}%
\begin{subfigure}{.5\textwidth}
  \centering
\includegraphics[width=1.0\linewidth]{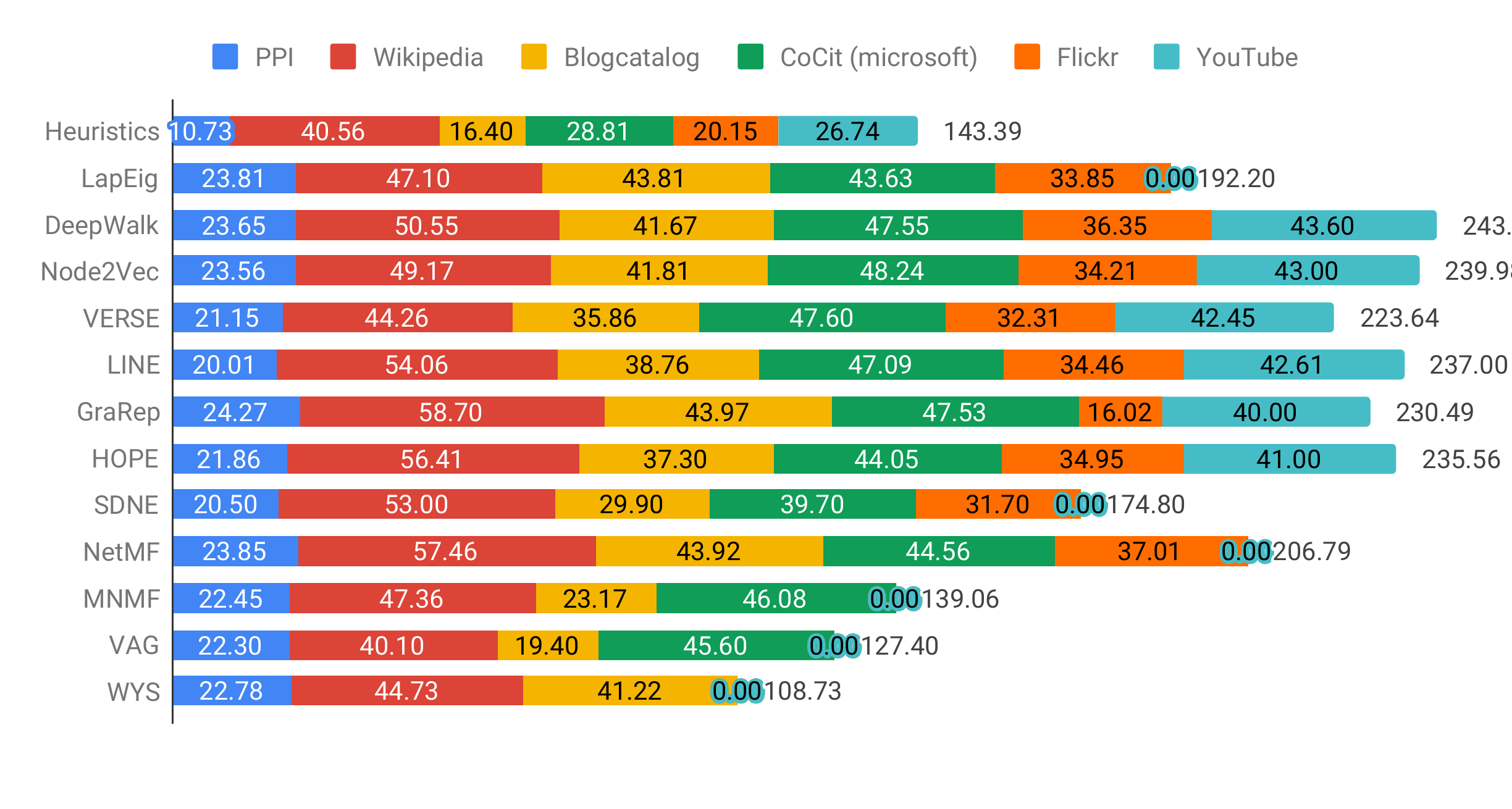}
  \caption{Datasets with more labels}
  \label{fig:micro_more_label_nl}
\end{subfigure}
\caption{The node classification performance measured with Micro-f1 on train-test split of 50:50  with non-linear classifier.}
\label{fig:node_micro_nl}
\end{figure*}

\begin{figure}[t]
    \centering
    \includegraphics[width=1.0\linewidth]{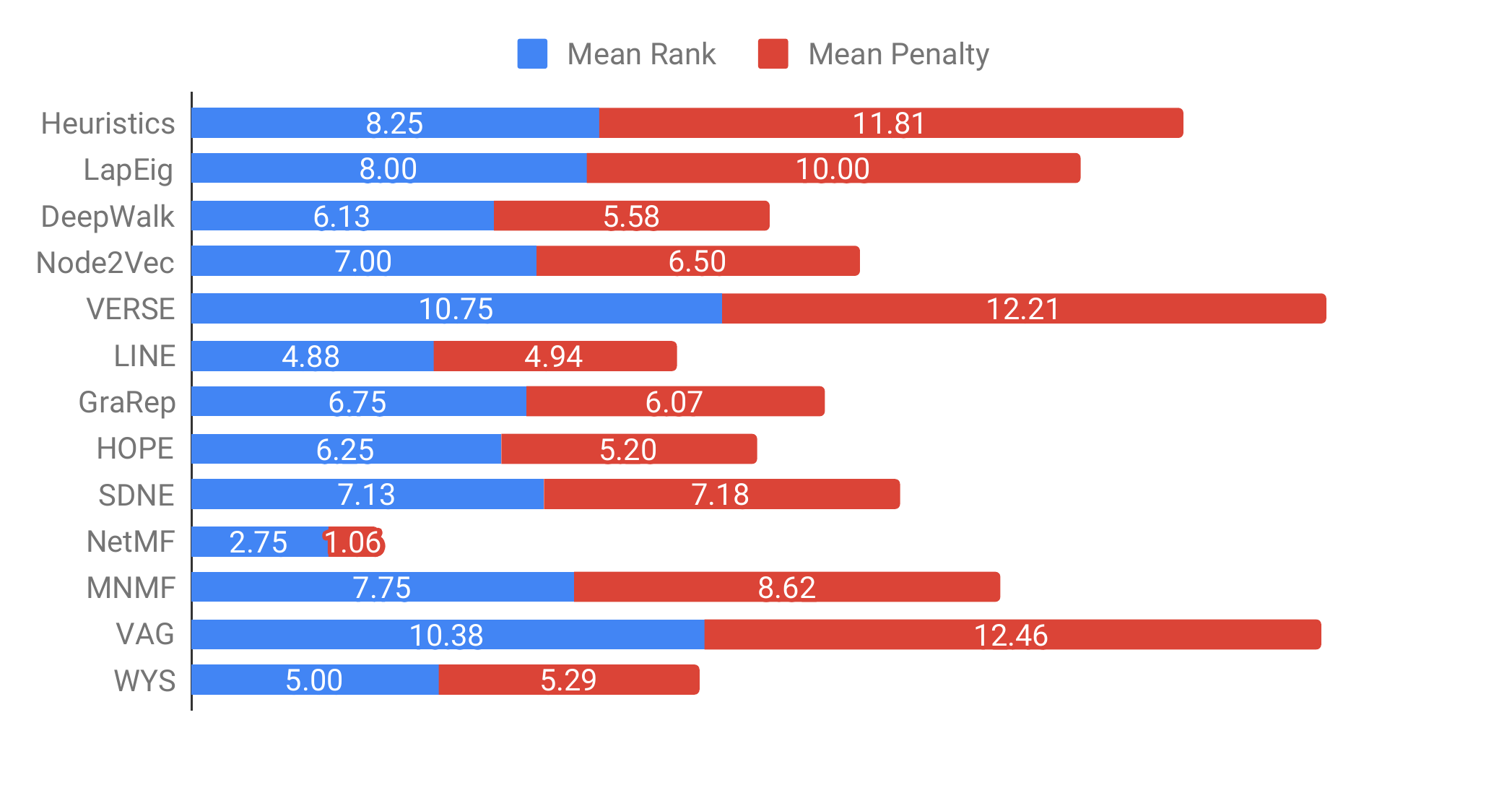}
    \caption{Mean Rank and Mean Penalty -- on all datasets where all methods finish execution -- for node classification with selected performance metric as Micro-f1 and Logistic Regression classifier.}
    \vspace{-1em}
    \label{fig:mr_op_nc_all}
\end{figure}

\begin{figure}[t]
    \centering
    \includegraphics[width=1.0\linewidth]{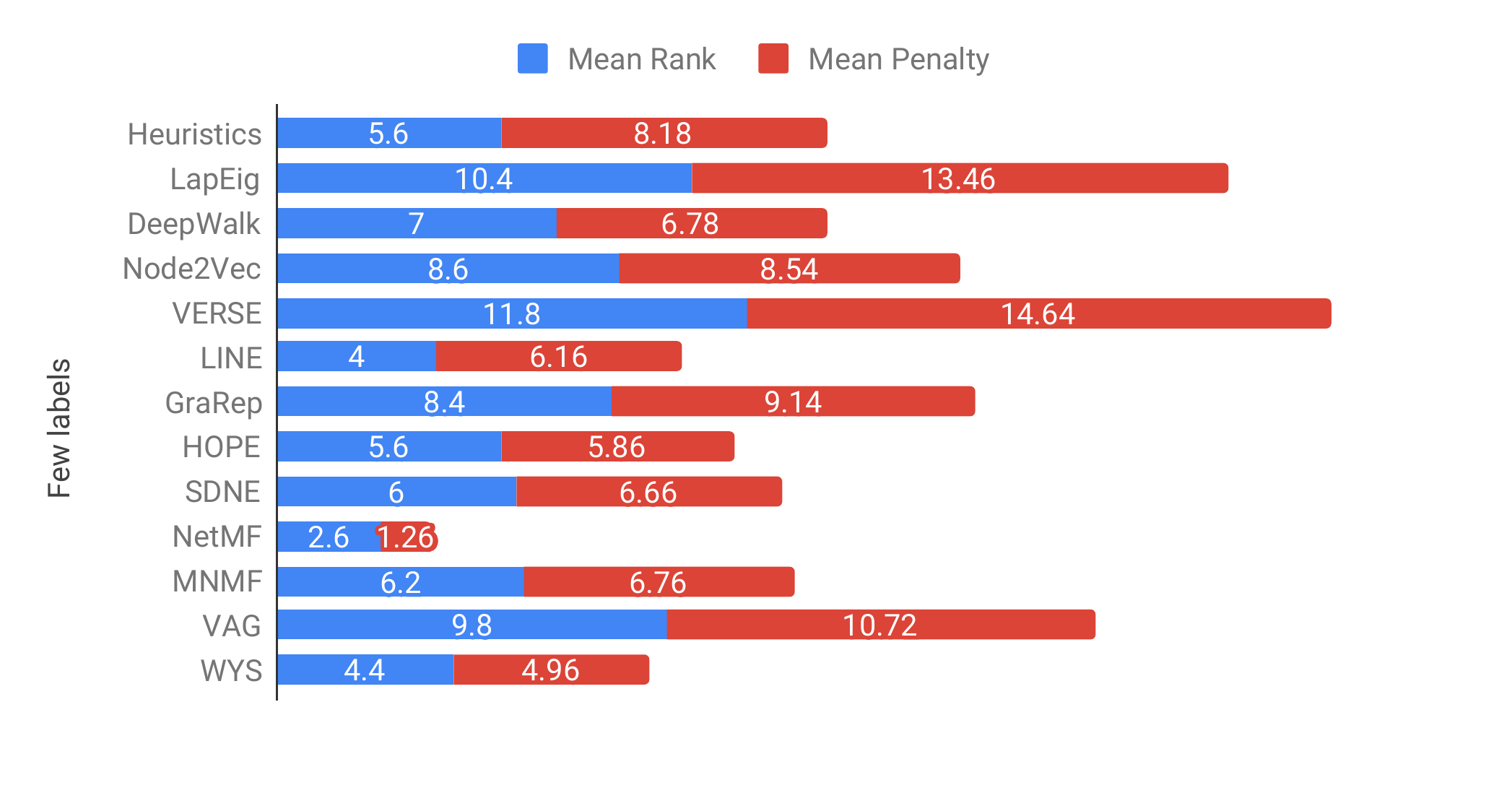}
    \caption{The Mean Rank and Mean Penalty -- on  datasets with few labels -- for node classification with selected performance metric as Micro-f1 and Logistic Regression classifier.}
    \vspace{-1em}
    \label{fig:mr_op_nc}
\end{figure}

\begin{figure*}[t]
\centering
\begin{subfigure}{.48\textwidth}
\includegraphics[width=1.0\linewidth]{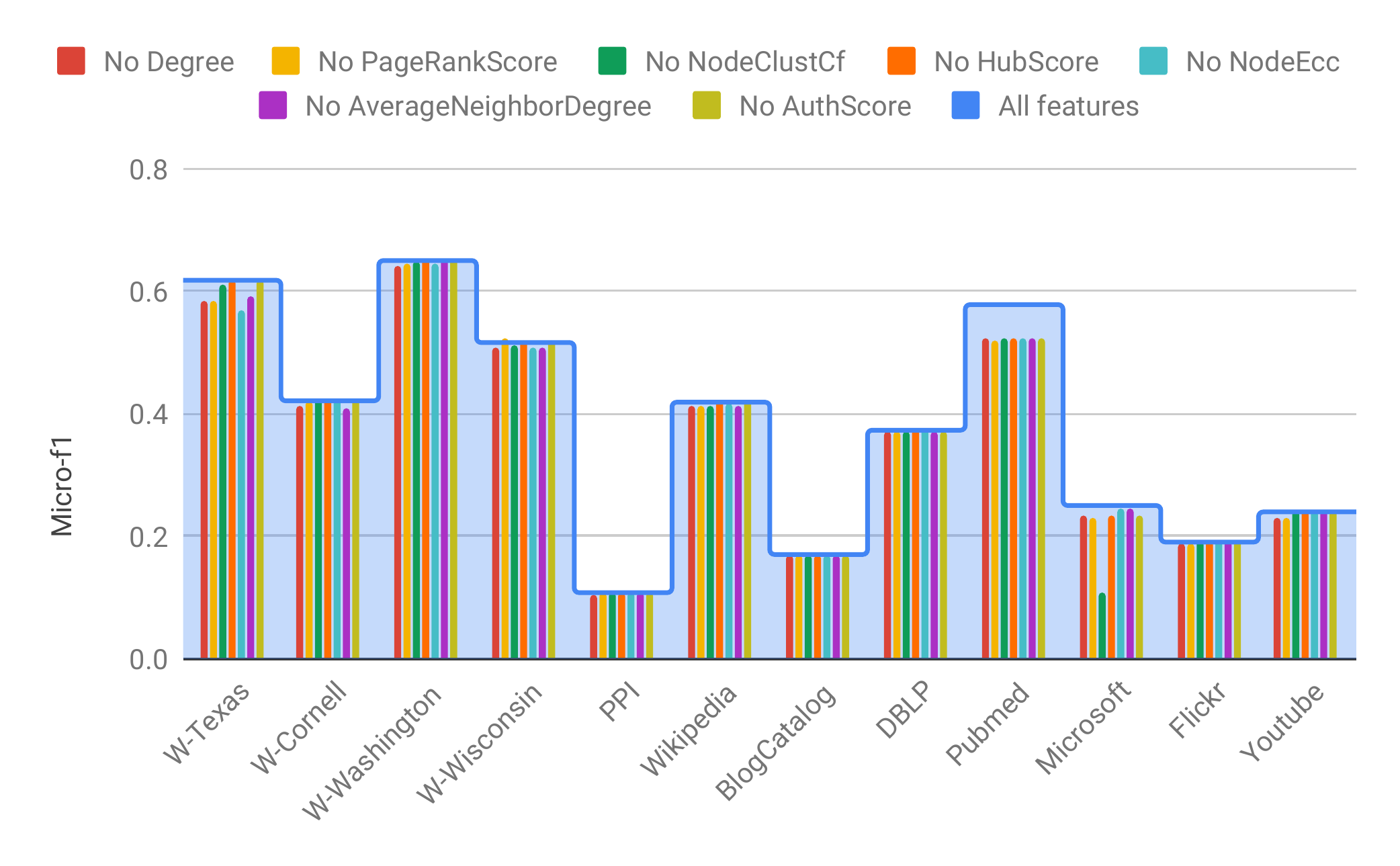}
\caption{Linear classifier : Logistic Regression}
\label{fig:nc_heu_ablation_lr}
\end{subfigure}%
\begin{subfigure}{.48\textwidth}
\includegraphics[width=1.0\linewidth]{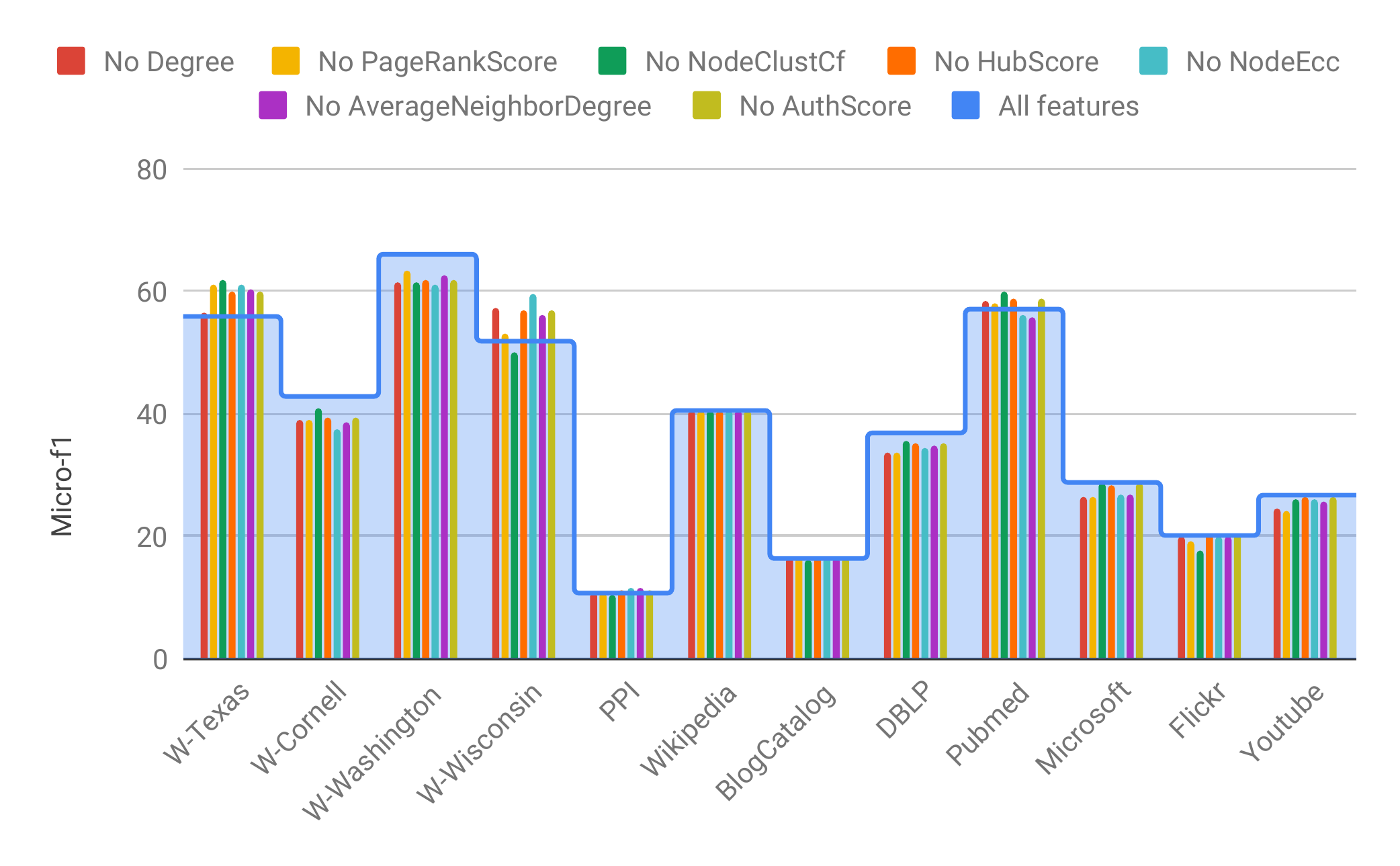}
\caption{Nonlinear classifier : EigenPro}
\label{fig:nc_heu_ablation}
\end{subfigure}
\caption{Feature study on the node classification for node classification heuristics.}
\end{figure*}

\subsection{Link Prediction}
\label{lp_results}
The link prediction performance of 12 embedding methods measured in terms of AUROC and AUPR on 15 datasets is shown in Figure \ref{fig:LP-AUC} and Figure \ref{fig:LP-AUPR}. The {\it Overall} (or aggregate) performance of an embedding method on all the datasets is also shown at the end of the horizontal bar of each method in Figure \ref{fig:LP-AUC} and Figure \ref{fig:LP-AUPR}. We represent {\it Overall} score as the sum of scores (AUROC or AUPR)  of the method on all the datasets. The {\it Mean Rank} and {\it Mean Penalty} of embedding methods -- on datasets for which all methods run to completion on our system --  is shown in Figure \ref{fig:mr_op_lp}. We also provide the tabulated results in Tables \ref{table:lp_auroc} and \ref{table:lp_aupr}. As mentioned in Section \ref{network_methods}, we tune the hyper-parameters of each embedding method and report the best average AUROC scores and average AUPR scores across 5 folds. In the case of WebKB datasets, we evaluate the methods on embedding dimensions 64, 128. We perform the link prediction task with both normalized and unnormalized embeddings and report the best performance.


We make the following observations:

\begin{itemize}[leftmargin=*]
    
    \item \textbf{Effectiveness of MNMF for Link Prediction}: We observe that MNMF achieves the highest overall link prediction performance in terms of best average AUROC and AUPR scores as compared to other methods.  The competitive performance of MNMF on link prediction could be credited to the community information imbibed into the node embeddings generation. The {\it Mean Rank} and {\it Mean Penalty} is lowest for MNMF which also suggest MNMF as a competitive baseline for Link Prediction. MNMF achieves the first rank for 7 out of 15 datasets. The small value of {\it Mean Penalty} suggests that even when MNMF is not the top-ranked method for a particular dataset, MNMF's performance is closest to that of the top-ranked method on that dataset.  However, MNMF does not completely outperform other methods on all the datasets. For instance, on the Wiki-Vote and Pubmed dataset, WYS achieves the best average AUROC scores while on Microsoft dataset, GraRep achieves the best average AUROC score. 
    In Figure \ref{fig:lp_auc_large} and Figure \ref{fig:lp_aupr_large}, we see that among the more scalable methods, LINE achieves the highest overall link prediction performance followed by DeepWalk in terms of both AUROC and AUPR scores.
    Note that MNMF did not scale for the datasets with $\ge$5M edges on a modern machine with 500 GB RAM and 28 cores. However, the scalability issue of non-negative matrix factorization based methods can be addressed by adopting modern ideas \cite{moon2019pl, liang2018mile} (outside the scope of this study).

    \item \textbf{Performance of Heuristic Baseline: } We observe that the Link Prediction Heuristics baseline -- described in section \ref{lp_heuristics} -- is both efficient and effective. We see that Link Prediction Heuristics baselines' overall performance is better than that of Laplacian Eigenmaps and SDNE and competitive to that of Node2vec, HOPE, Verse, LINE. The {\it Mean Penalty} of Link Prediction Heuristics is also close to other embedding methods. On the largest dataset YouTube, Link Prediction Heuristics achieve an AUROC of 96.2\% which is close to the best performing Verse with AUROC of 97.6\%. As compared to the most competitive baseline MNMF, the heuristics baseline outperforms MNMF on Wikipedia, Blogcatalog datasets. We also observe that the heuristics baseline performance is competitive against several methods on the directed datasets too even though the chosen similarity-based metrics in heuristics baseline treat the underlying graph as undirected.

\textit{Feature study on the Heuristic Baseline:}
We study the importance of the individual feature in the heuristics by analyzing the impact of the feature removal on link prediction. The results are reported in Figure \ref{fig:lp_heu_ablation}. The blue line on the top of the columns in Figure \ref{fig:lp_heu_ablation}  corresponds to AUROC scores achieved  with the proposed link prediction heuristic. In the feature study on the link prediction heuristics, we see that the removal of preferential attachment (PA) feature results in consistent drop in AUROC scores. We find that the removal of PA feature results in statistically significant  drop  at significance level of 0.05 with  paired t-test. The removal of rest of the features both in link prediction heuristics did not result in significant drop in the downstream performance.


    \item \textbf{Impact of Evaluation Strategy:} As described in section \ref{lp_setup}, the presence of a link between two nodes can be predicted with either the Logistic Regression classifier (treating the embeddings as features) or the dot product between the node embeddings. We compare the performance of both evaluation strategies on each embedding method over all datasets using the differences in the average AUROC scores.  A positive difference implies link prediction performance with classifier is better than that of the dot product. The results are presented as box-plot in Figure \ref{fig:classi_vs_dot_product}. Paired t-test suggests the positive difference is statistically significant for all methods, except for Verse and WYS, with a significance level of 0.05. Hence, the use of classifier over dot product provides significant predictive performance gain on the task of link prediction. We also investigate the changes in the ranking of embedding methods based on overall average AUROC scores when predictions are performed with classifier rather than dot product. The methods were ranked based on overall average AUROC scores and we considered only those datasets on which all methods complete execution. We observed that the rank of NetMF in the ranking generated with dot product was 10 while its rank in the ranking generated with classifier improved to 3. Since the best link prediction performance for the majority of the embedding methods was achieved with classifier, we believe the superiority of the embedding methods based on link prediction task should be asserted by leveraging the classifier.

    As mentioned in section \ref{lp_setup}, we lever binary functions: Hadamard, Concatenation, and L2 to generate the edge embedding. In figure \ref{fig:lp_features}, we present which binary function achieved the best average AUROC score for an embedding method on a particular dataset. We see that the binary function Hadamard resulted in achieving a maximum number of best average AUROC scores. However, there is no single winner in terms of choice of binary functions. 

    \item \textbf{Impact of context embeddings:} We study the impact of context embeddings on directed datasets for the link prediction task. We consider only those embedding methods which generate both node and context embeddings for this study. We compare the impact of using node + context embeddings over using only node embeddings with the help of differences in AUROC scores. The results are detailed in Figure \ref{fig:cxt_no_cxt}. A positive difference implies the use of context embeddings helps in link prediction. We see that levering node + context embeddings improve the link prediction performance of LINE, HOPE, and WYS. For MNMF, use of context embeddings does not improve the link prediction performance as in MNMF the community information -- crucial for link prediction -- is incorporated in the node embeddings. In the case of GraRep, we find that the node embeddings encapsulate high-order information and, hence, levering context does not help improve the performance. We find that the results of DeepWalk and Node2Vec on directed datasets are significantly lower, so in order to have a fair comparison with other embedding methods, we treated the directed datasets as undirected for DeepWalk and Node2Vec. The median of the box plot of DeepWalk and Node2Vec is close to zero due to this treatment.

\item \textbf{Robustness of embedding methods:}
  In link prediction, we compute the average AUROC score and average AUROC standard error of an embedding method over 5 folds of a selected dataset.  The computed average AUROC standard error corresponds to the robustness of that embedding method on the selected dataset -- as larger values of standard error corresponds to large variance in AUROC scores across 5 folds. In Figure \ref{fig:stddev}, we report the distribution of average AUROC standard error of each embedding method over all datasets. We observed large variance in average AUROC standard error scores over WebKB datasets and show the results for WebKB datasets in Figure \ref{fig:lp_std_webkbonly} while we show the results for other datasets in Figure \ref{fig:lp_std_webkbno}.  Interestingly, even on WebKB datasets, the variance in average AUROC standard error scores is low for MNMF method. From Figure \ref{fig:lp_std_webkbno}, we observe that the median of box-plots of majority of the methods is closer to zero.

\item \textbf{Impact of embedding dimension:}
We study the impact of embedding size for all the embedding methods on the link prediction. Specifically, we compare the performances of 64 dimensional embedding with 128 dimensional embedding. The improvement -- quantified in terms of performance difference -- obtained with 128 dimensional embedding over 64 dimensional embedding is reported in Figure \ref{fig:lp_dim}. The box-plot represents the distribution of differences in AUROC scores between 128 dimensional embedding and 64 dimensional embedding for each method on all datasets. In link prediction, we observe a statistically significant improvement at significance level 0.05 with the 128 dimensional embedding for Laplacian Eigmaps, GraREP, HOPE, NetMF and MNMF methods.

\item \textbf{Impact of embedding normalization:}
We study the impact of L2 normalization of the embeddings on the link prediction performance. The  comparison results are shown in Figure \ref{fig:lp_norm} where Figure \ref{fig:lp_l2norm}  and Figure \ref{fig:auc_lp_l2norm_lr} shows the comparison results when link prediction is performed through classifier and dot-product, respectively.  The box plot represents the distribution of differences of AUROC between normalized and unnormalized embeddings on link prediction task. The positive difference implies L2 normalization results in better downstream performance. 
When link prediction is performed through classifier, the negative difference is statistically significant for VERSE and GraREP at significance level of 0.05 with paired t-test. However, surprisingly the difference in performance with respect to normalization of embeddings is not statistically significantly for rest of the methods for link prediction. 
When link prediction is performed through dot-product, the normalization of embedding results in statistically significant improvement for Node2vec and Verse, while not performing normalization of embedding results in statistically significant improvement HOPE, NetMF and WYS.

\end{itemize}

%% file: NC_results.tex
\subsection{Node Classification}
\label{nc_results}

\begin{figure}[t]
    \centering
    \includegraphics[width=0.85\linewidth]{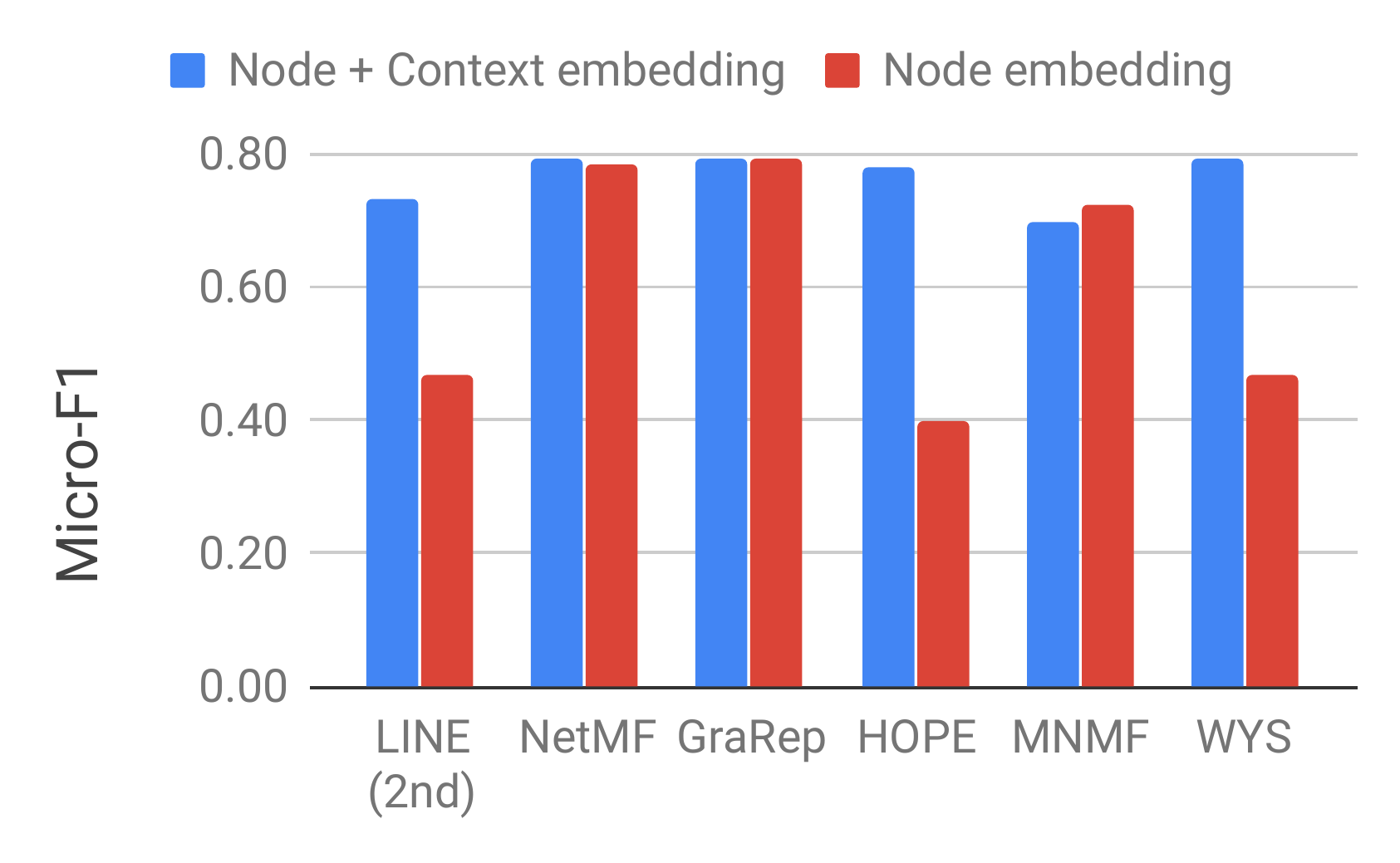}
    \caption{Node Classification on Directed Dataset (PubMed) with/without concatenation of Node embeddings and Context Embeddings (128 dimensions).}
    \label{fig:nc:pubmed:wc_woc}
    \vspace{-1em}
\end{figure}

\begin{figure}[t]
    \centering
     \includegraphics[width=\linewidth]{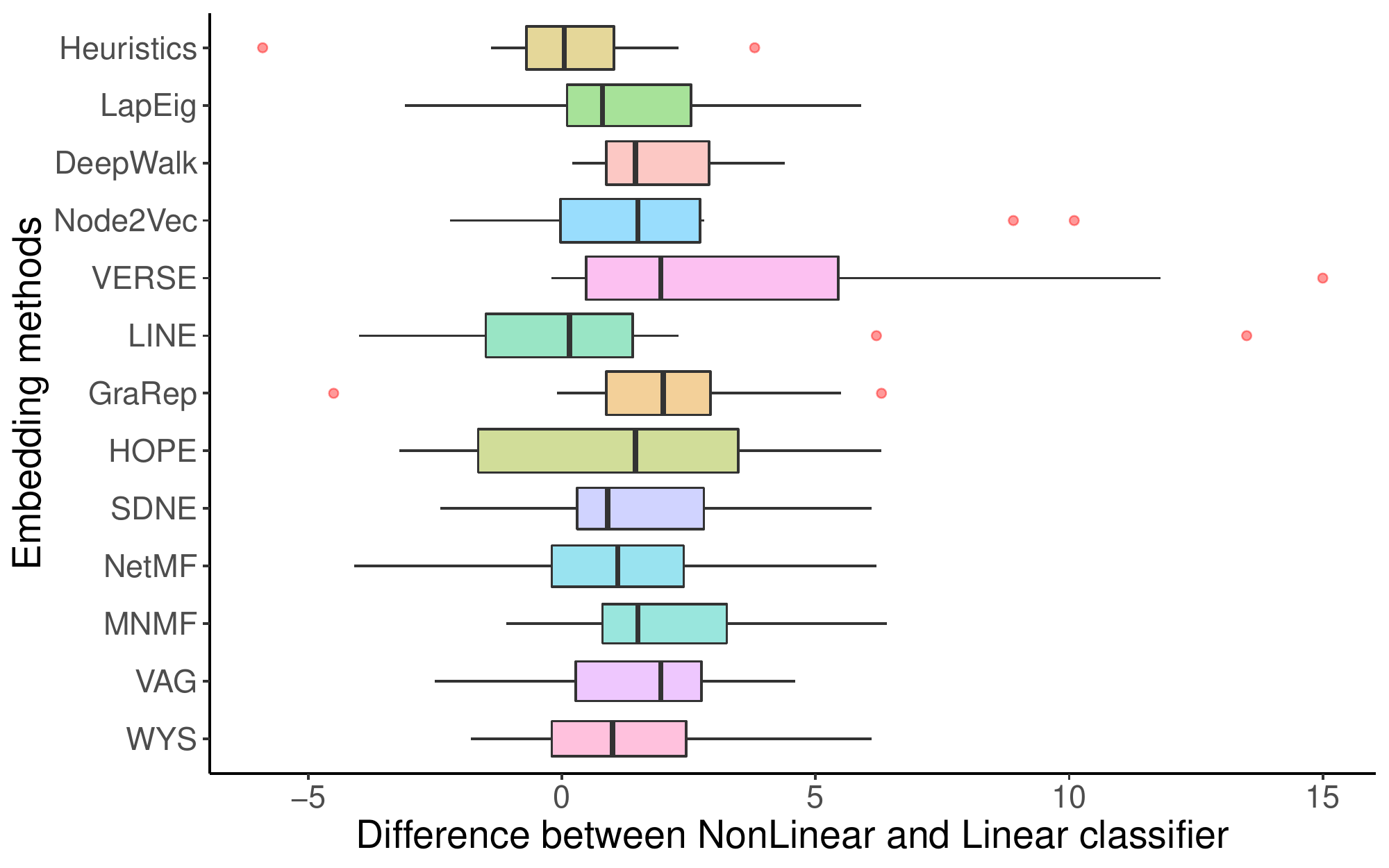}
    \caption{Box-plot represents the distribution of differences between non-linear and linear classifier on all the datasets. }
    \label{fig:linear_vs_nonlinear}
    \vspace{-1em}
\end{figure}

\begin{figure*}[t]
\centering
\begin{subfigure}{.48\textwidth}
\includegraphics[width=0.95\linewidth]{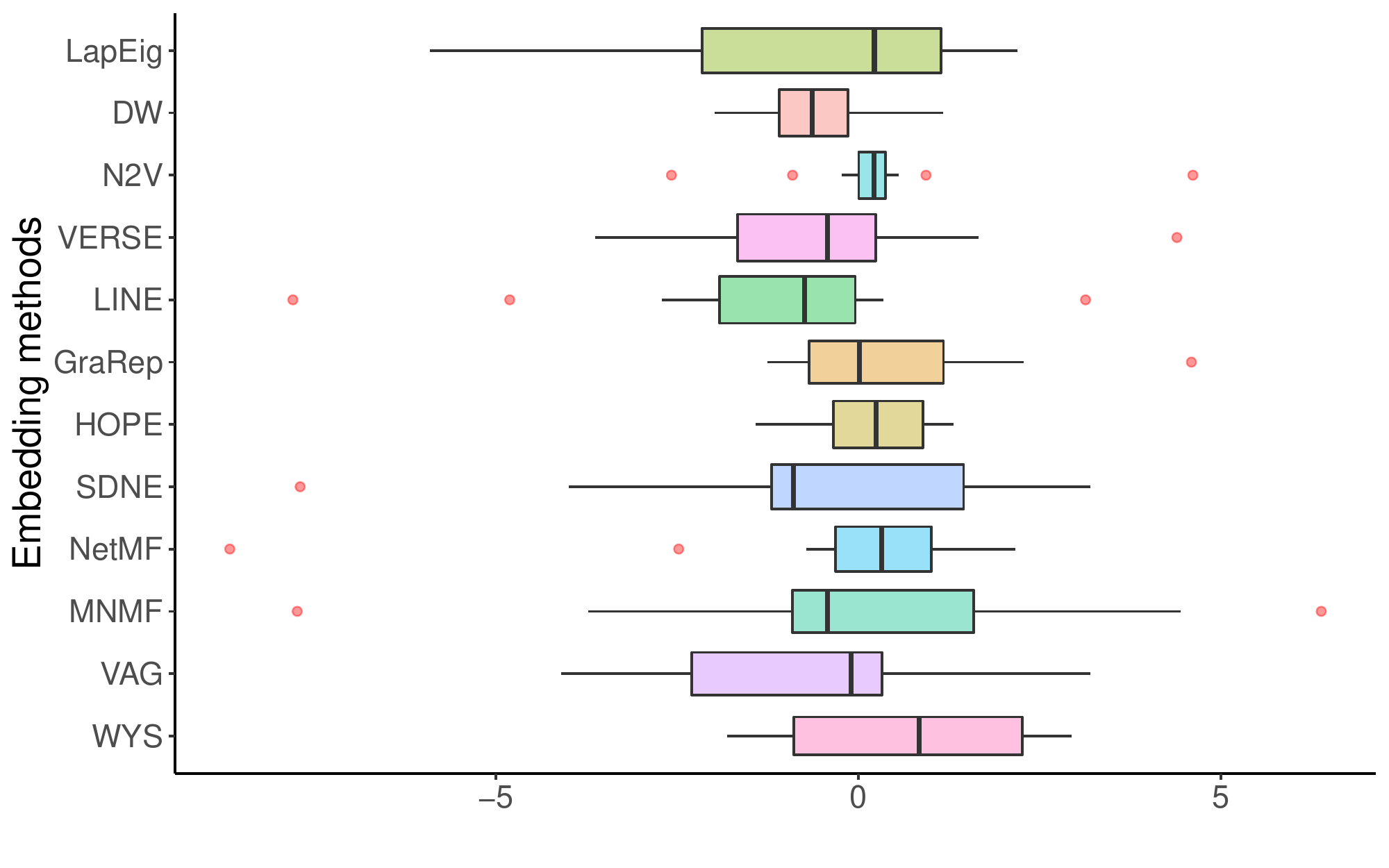}
\caption{Linear classifier : Logistic Regression}
\label{fig:mr_nc_dim_lr}
\end{subfigure}%
\begin{subfigure}{.48\textwidth}
\includegraphics[width=0.95\linewidth]{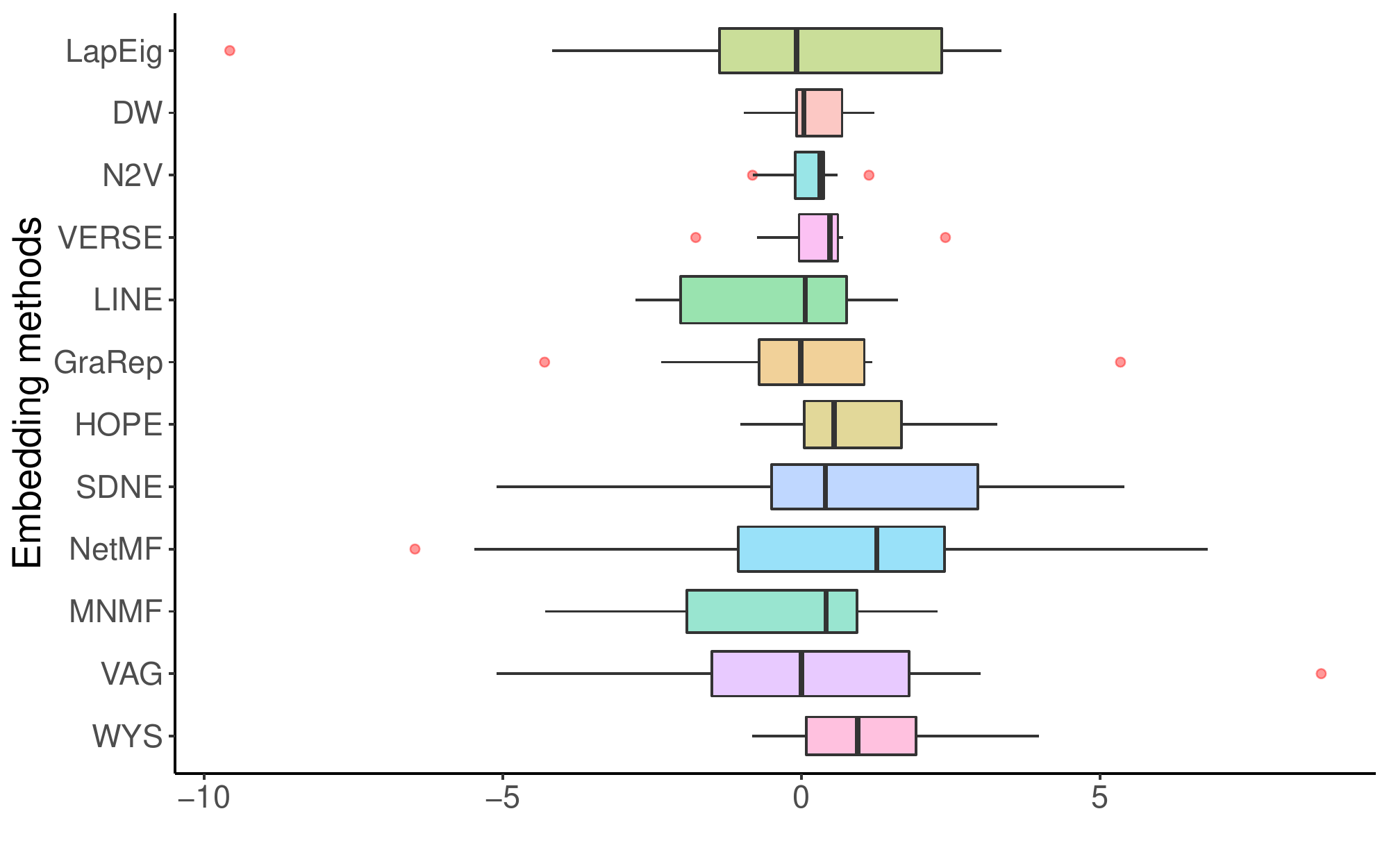}
  \caption{Nonlinear classifier : EigenPro}
  
  \label{fig:nc_dim_eig}
\end{subfigure}
\caption{The box-plot represents the distribution of the differences in Micro-f1 scores between 128 dimensional embedding and 64 dimensional embedding for each method on all datasets}
\label{fig:dimensions}
\label{fig:nc_dim}
\end{figure*}

\begin{figure*}[t]
\centering
\begin{subfigure}{.48\textwidth}
\includegraphics[width=1.0\linewidth]{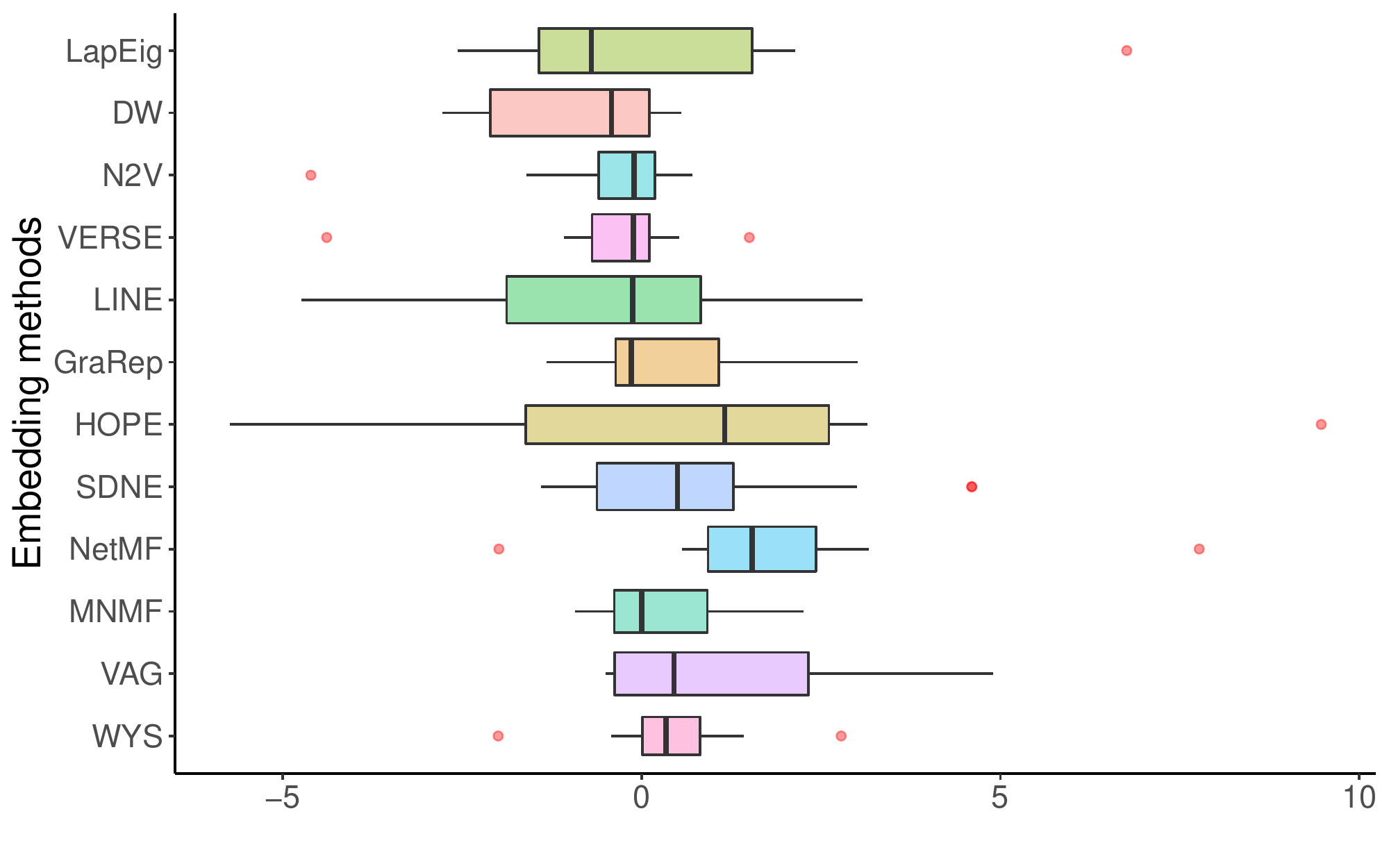}
\caption{Linear classifier : Logistic Regression}
\label{fig:mr_nc_l2norm_lr}
\end{subfigure}%
\begin{subfigure}{.48\textwidth}
\includegraphics[width=1.0\linewidth]{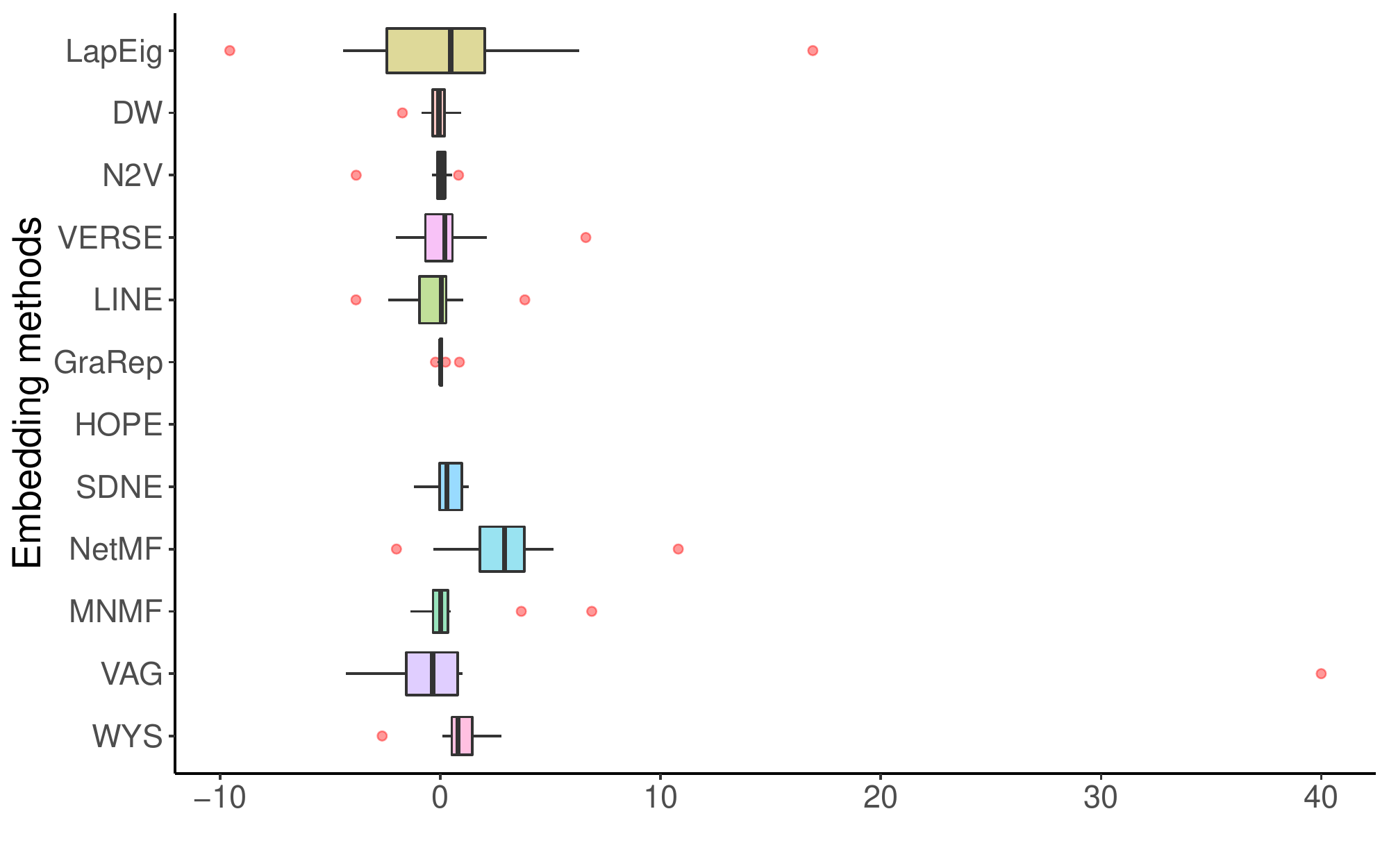}
\caption{Nonlinear classifier : EigenPro}
\label{fig:nc_l2norm}
\end{subfigure}
\caption{Impact of embedding normalization on node classification performance.}
\label{fig:nc_norm}
\end{figure*}

\begin{figure*}[t]
\centering
\begin{subfigure}{.5\textwidth}
 \includegraphics[width=1.0\linewidth]{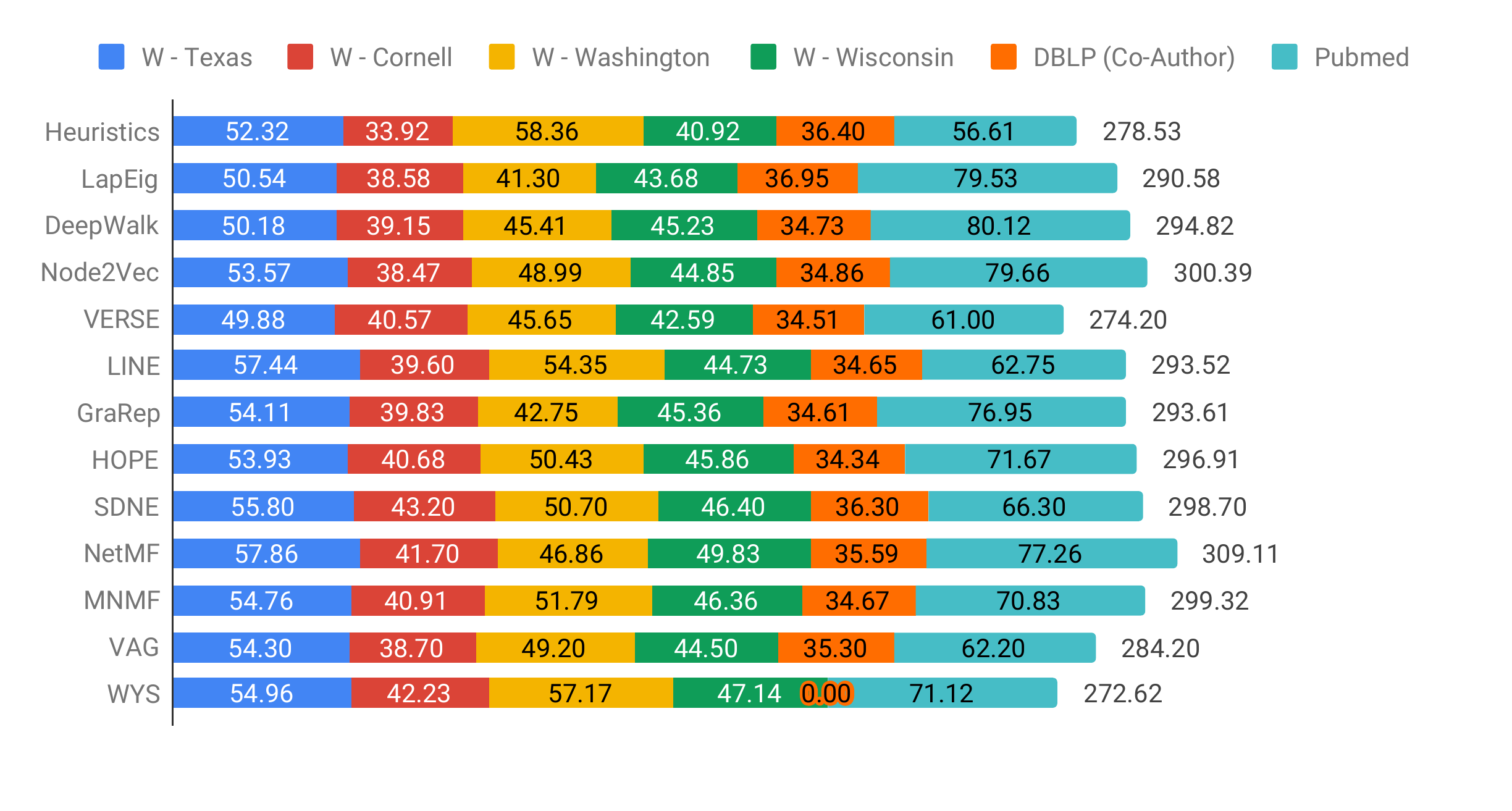}
  \caption{Datasets with few labels}
  \label{fig:micro_few_label_10}
  
\end{subfigure}%
\begin{subfigure}{.5\textwidth}
\includegraphics[width=1.0\linewidth]{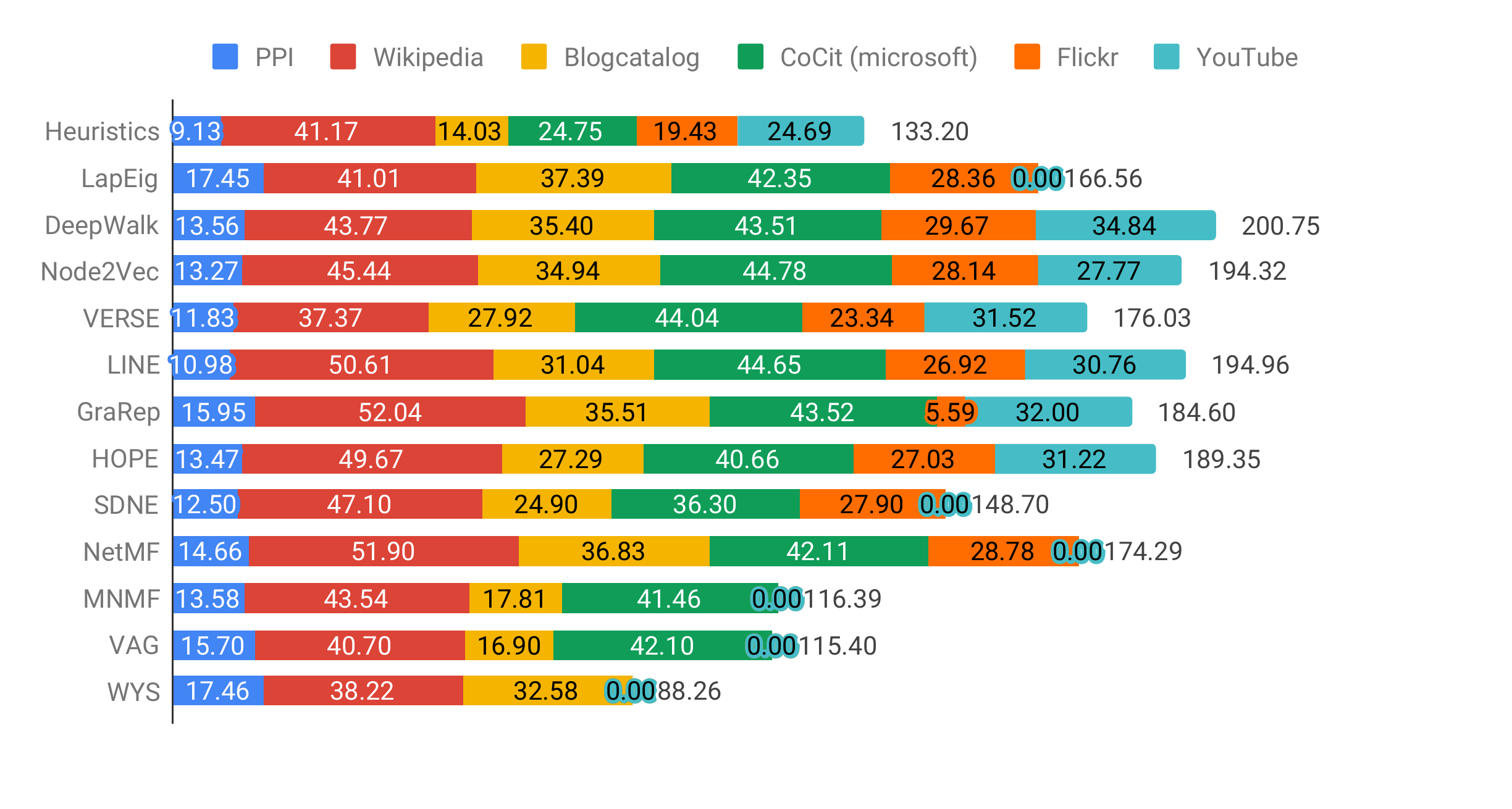}
  \caption{Datasets with more labels}
  \label{fig:micro_more_label_10}
\end{subfigure}
\caption{The node classification performance measured  with Micro-f1 on train-test split of 10:90  with Logistic Regression. For each method, the number at the end of bar represent the summation of the Micro-f1 values across the datasets. }
\vspace{-1em}
\label{fig:node_micro_lr_10}
\end{figure*}

\begin{figure*}[t]
\centering
\begin{subfigure}{.5\textwidth}
\includegraphics[width=1.0\linewidth]{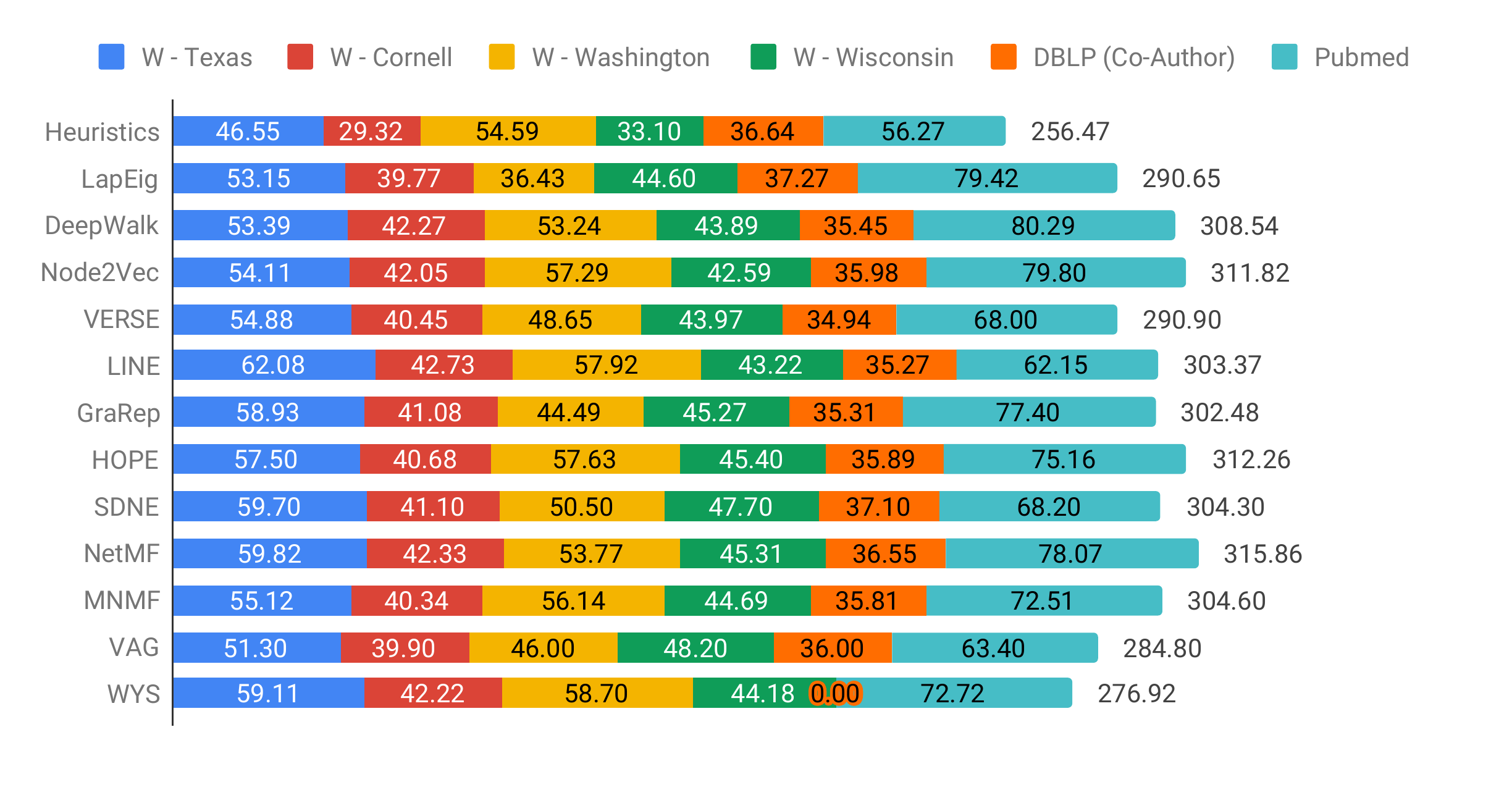}
  \caption{Datasets with few labels}
  \label{fig:micro_few_label_nl_10}
\end{subfigure}%
\begin{subfigure}{.5\textwidth}
\includegraphics[width=1.0\linewidth]{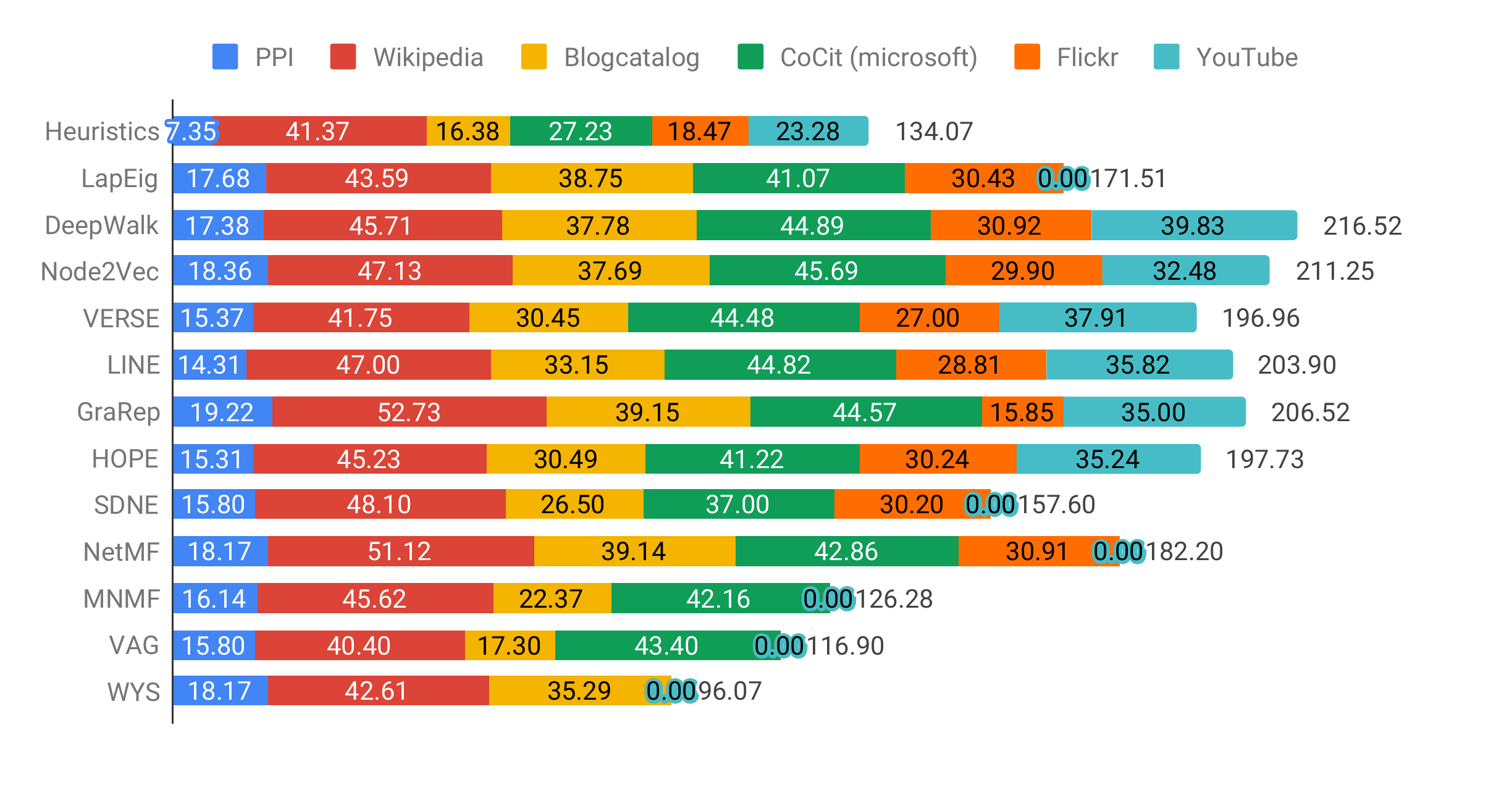}
  \caption{Datasets with more labels}
  \label{fig:micro_more_label_nl_10}
\end{subfigure}
\caption{The node classification performance measured with Micro-f1 on train-test split of 10:90  with non-linear classifier.}
\label{fig:node_micro_eig_10}
\end{figure*}

\begin{figure*}[!htb]
\centering
\begin{subfigure}{.48\textwidth}
\includegraphics[width=1.0\linewidth]{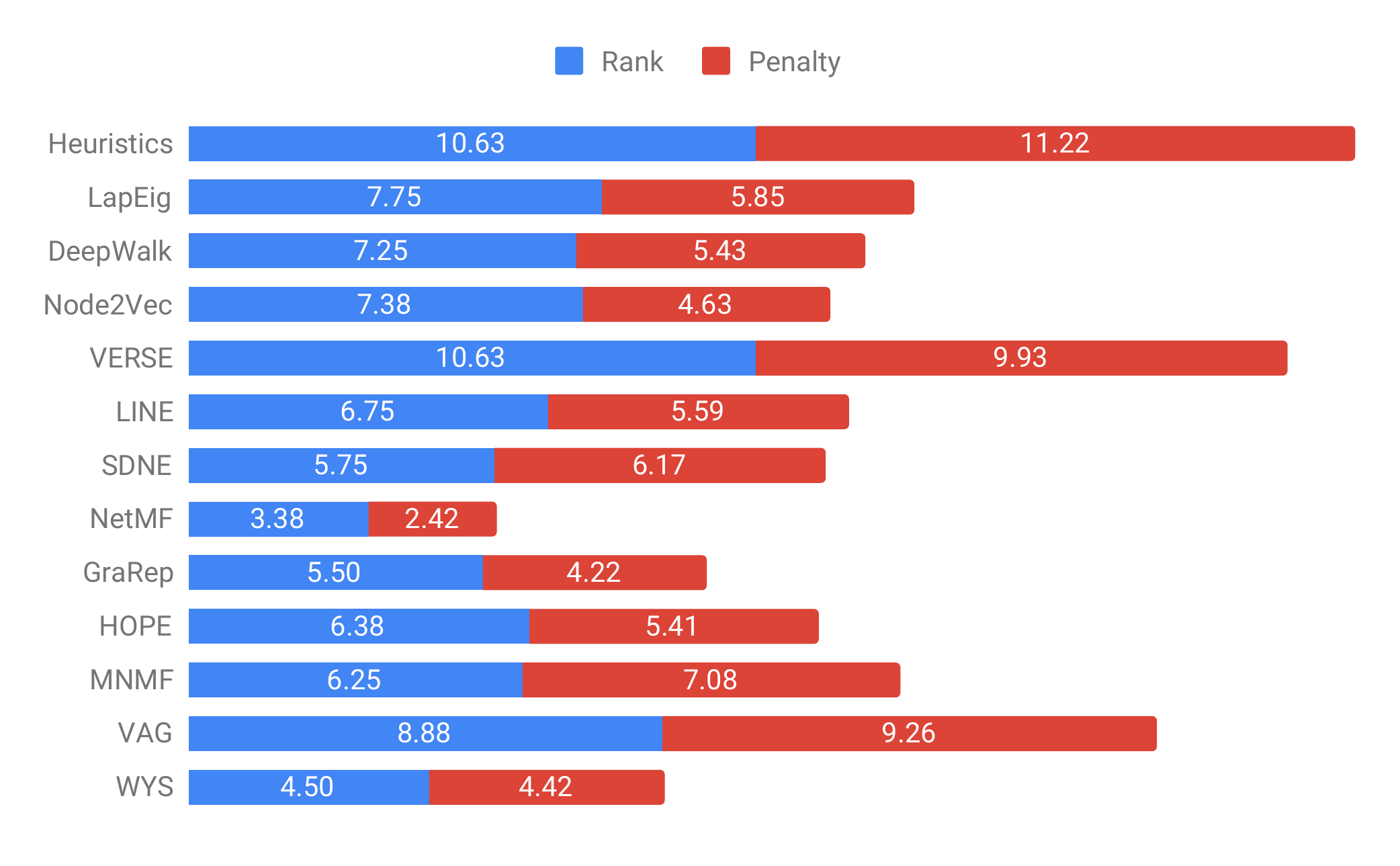}
\caption{On all datasets where all methods finish execution}
\label{fig:mr_op_nc_all_10}
\end{subfigure}%
\begin{subfigure}{.48\textwidth}
\includegraphics[width=1.0\linewidth]{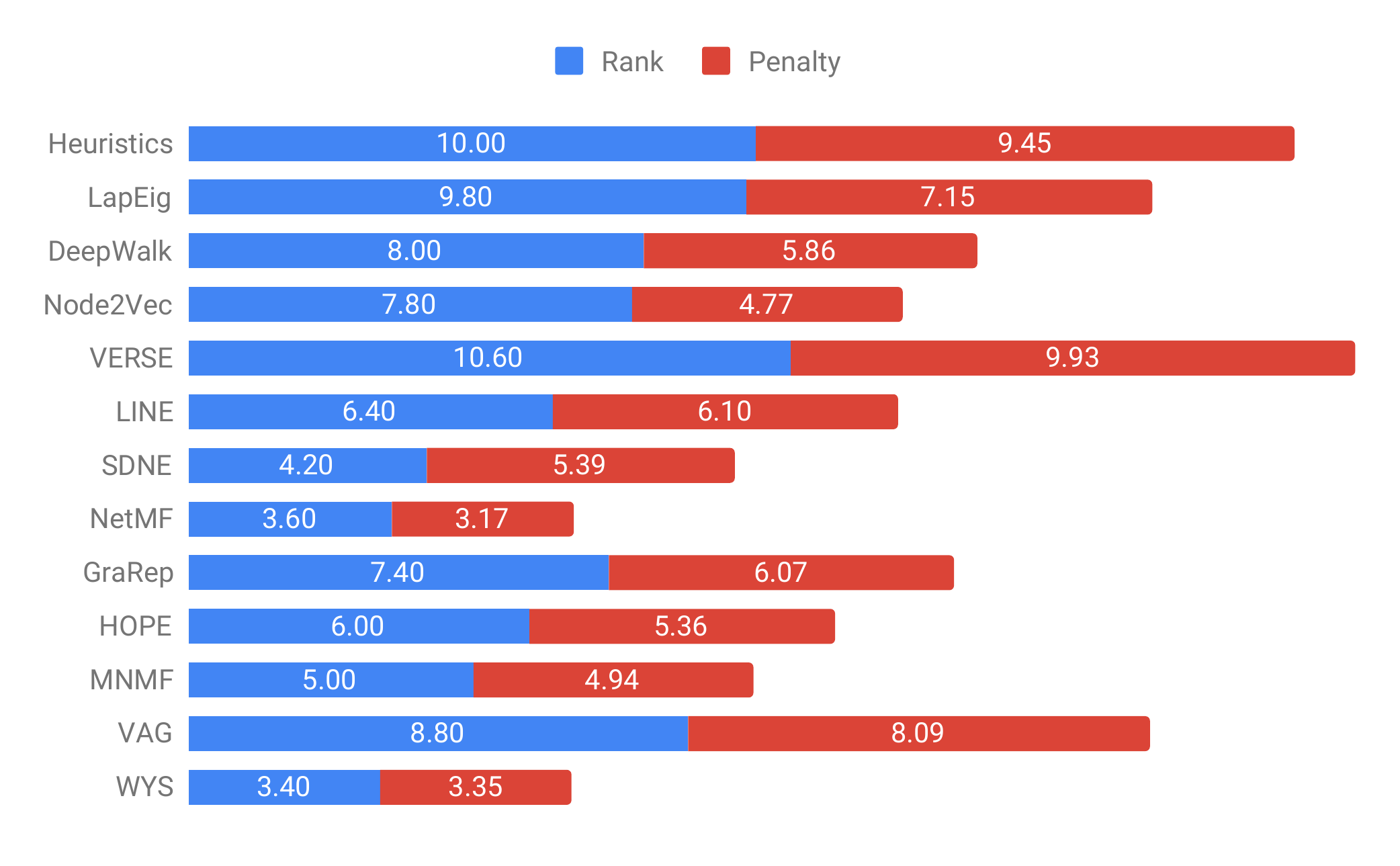}

\caption{On datasets with few labels }
\label{fig:mr_op_nc_10_fc}
\end{subfigure}
\caption{The Mean Rank and Mean Penalty -- on  datasets with few labels -- for node classification with selected performance metric as Micro-f1 on train-test split of 10:90 and logistic regression as the classifier.}
\label{fig:mr_op_nc_10_all}

\end{figure*}

The node classification performance of 12 embedding methods measured in terms of Micro-f1 scores on 15 datasets with train-test split of 50:50 is reported in Figure \ref{fig:node_micro} and Figure \ref{fig:node_micro_nl}. The {\it overall} performance of an embedding method on all the datasets is shown at the end of the horizontal bar of each method in Figure \ref{fig:node_micro} and Figure \ref{fig:node_micro_nl} and represents the sum of scores (Micro-f1) of the method on the datasets. The {\it Mean Rank} and {\it Mean Penalty} of embedding methods -- on the datasets for which all methods run to completion on our system -- is shown in Figure \ref{fig:mr_op_nc_all}. We also report the {\it Mean Rank} and {\it Mean Penalty} of embedding methods -- on datasets with few labels -- in Figure \ref{fig:mr_op_nc}.  We tune the hyper-parameters of each embedding method -- mentioned in section \ref{section:models} -- and report the best Micro-f1 score. In the case of WebKB datasets, we evaluate the methods on generated embedding with dimensions 64, 128. We perform the node classification with both normalized and unnormalized embeddings and report the best performance. We also provide tabulated results in Tables \ref{table:nc-micro} and \ref{table:nc-macro}. We make the following observations.
\begin{itemize}[leftmargin=*]

\item \textbf{Effectiveness of NetMF for node classification:} 
We observe that NetMF achieves the highest overall node classification performance in terms of best Micro-f1 scores using both linear and non-linear classifiers. 
From Figure \ref{fig:mr_op_nc_all}, we see that the {\it Mean Rank} and {\it Mean Penalty} is lowest for NetMF which suggest NetMF as the strongest overall method for node classification.  NetMF achieves low {\it Mean Rank} suggesting NetMF is among the top-ranked methods on the evaluated datasets.  The smallest value of {\it Mean Penalty} suggests that even when NetMF is not the top-ranked method for a particular dataset, NetMF's performance is closest to that of the top-ranked method for that dataset. However, it does not entirely outperform other methods on all the datasets. LINE, DeepWalk, and Node2Vec are also competitive baselines for the task of node classification as their overall performance is closest to that of NetMF. The performance of GraRep on datasets with more labels is comparable with other methods when we exclude the Flickr dataset. However, the reported results for GraRep on the Flickr dataset are with embeddings of dimension 64. Embedding dimensions 128 and 256 for GraRep resulted in memory errors on a modern machine with 500GB RAM and 28 cores. Note that NetMF did not scale for the YouTube dataset. While scalability is currently outside the scope of our study, the scalability of such methods is under active development (we refer the interested reader elsewhere \cite{qiu2019netsmf, liang2018mile}).

\item \textbf{Laplacian Eigenmaps Performance:} We observe that the Laplacian Eigenmaps method achieves competitive Micro-f1 scores on several datasets. For instance, on Blogcatalog dataset with 39 labels, Laplacian Eigenmaps method achieves the best Micro-f1 score of 42.1\% while on the Pubmed dataset, the Laplacian Eigenmaps methods outperform all other embedding methods with 81.7\% Micro-f1. With a non-linear classifier, Laplacian Eigenmaps achieves the second-best performance on the PPI dataset with 23.8\% Micro-f1. We observe from Figure \ref{fig:mr_op_nc_all} that the {\it Mean Penalty} of Laplacian Eigenmaps is also closest to other embedding methods, namely, Verse, MNMF, VAG. On the PPI and Flickr datasets, Laplacian Eigenmaps baselines' Micro-f1 is close to the best Micro-f1.  The observed results for Laplacian Eigenmaps on evaluated datasets are better than the reported results \cite{goyal2018graph, grover2016node2vec} for both node classification and link prediction tasks. This improvement in the performance of Laplacian Eigenmaps is due to the hyper-parameter tuning of parameters of logistic regression classifier. 
    
\item \textbf{Node Classification Heuristic:} We observe from Figure \ref{fig:micro_few_label} and Figure \ref{fig:micro_few_label_nl} that the node classification heuristic baseline is competitive against other embedding methods on datasets with fewer labels (up to 5 labels) as its overall score is better than {\it many of the methods}. This observation can also be verified from Figure \ref{fig:mr_op_nc} as both {\it Mean Rank} and {\it Mean Penalty} of node classification heuristics baseline is better than {\it many of the methods}. 
However, as the number of labels in the datasets increases ($>$5 labels), we observe that the Micro-f1 scores of node heuristics baseline decrease drastically. The decrease in overall performance reflects that the node heuristics features lack the discriminative power to classify multiple labels.


\textit{Feature study on the Heuristics baseline:}
 We study the importance of the individual feature in the node classification heuristics by analyzing the impact of the feature removal on the node classification performance. The results for node classification heuristic with logistic regression and EigenPro are reported in Figure \ref{fig:nc_heu_ablation_lr} and Figure \ref{fig:nc_heu_ablation}, respectively. The blue line on the top of the columns in the figures corresponds to Micro-f1 scores achieved  with the proposed node classification heuristics.  The removal of individual feature in node classification heuristics did not result in significant drop in the downstream performance. However, we see that the node classification heuristics' classification performance with both Logistic Regression and EigenPro is better than the ones achieved through the removal of individual features on most of the datasets.

\item \textbf{Context embeddings can improve performance:} We  see from Figure \ref{fig:nc:pubmed:wc_woc} that levering both node and context of Skip-gram based models results in significant improvement (up to 25\%) for most of the methods.
  On Pubmed dataset, we observe that the node classification performance of embedding methods like LINE (2\textsuperscript{nd} order), HOPE, and WYS was significantly lower than that of other methods. 
 The Micro-f1 scores of the embedding methods are shown in Fig. \ref{fig:nc:pubmed:wc_woc}. We found that the Pubmed dataset consists of around 80\% sink nodes.  
As a result, when the embedding methods based on Skip-gram model generate the node embeddings, the sink nodes are always considered as ``context" nodes and are never considered as ``source" nodes. Hence, the quality of node embeddings of sink nodes is of lower quality. In order to have a fair comparison, we concatenate both the node and context embeddings of the methods (whenever possible) and evaluate the performance on the concatenated embeddings.

\item \textbf{Impact of nonlinear classifier:} We study the impact of the nonlinear classifier on the node classification performance. The comparison results are shown in box plot \ref{fig:linear_vs_nonlinear}. The box plot represents the distribution of differences of Micro-f1 scores computed with the non-linear (EigenPro \cite{ma2017diving}) and linear classifiers (Logistic Regression). The positive difference implies that the results with non-linear classifier are better than linear classifier. For Verse, we see a 15\% absolute increase with the use of nonlinear classifier on the PubMed dataset. The positive difference is statistically significant (with paired t-test) for methods DeepWalk, Verse, SDNE, GraRep and MNMF with significance level 0.05.  It is worth pointing out that on the smaller datasets this gain is less evident while on, the larger datasets (more training data) the benefits of using a nonlinear classifier are much clearer.

\item \textbf{Impact of embedding dimension:}
We study the impact of embedding size for all the embedding methods on the node classification task. Specifically, we compare the performances of 64 dimensional embedding with 128 dimensional embedding. The improvement -- quantified in terms of performance difference -- obtained with 128 dimensional embedding over 64 dimensional embedding is reported in Figure \ref{fig:nc_dim}. The box-plot represents the distribution of differences in Micro-f1 scores between 128 dimensional embedding and 64 dimensional embedding for each method on all datasets. In node classification with linear classifier, none of the evaluated methods obtained a statistically significant difference at significance level 0.05. While in node classification  with non-linear classifier, the embedding method HOPE obtained a statistically significant positive difference -- at significance level 0.05 --  with 128 embedding dimension.

\item \textbf{Impact of embedding normalization:}
We study the impact of L2 normalization the embeddings for the node classification task. The  comparison results are shown in Figure \ref{fig:nc_norm}. The box plot represents the distribution of differences of Micro-f1 scores for embedding methods between normalized and unnormalized embeddings on node classification task.  The positive difference implies L2 normalization results in better downstream performance. In node classification with linear classifier, the positive difference is  statistically significantly for NetMF while the negative difference is  statistically significantly for DeepWalk at significance level 0.05 with paired t-test. While in node classification with non-linear classifier, the positive difference is  statistically significantly for NetMF method at significance level 0.05 with paired t-test.

\item \textbf{Node classification performance on 10:90 train: test split:} We report the node classification performance of all methods on all the evaluated datasets with 10:90 train:test split in with logistic regression classifier in Figure \ref{fig:node_micro_lr_10} and non-linear classifier EigenPro in Figure \ref{fig:node_micro_eig_10}. The Mean Rank and Mean Penalty of embedding methods – on the datasets for which all methods run to completion on our system – is shown in Figure \ref{fig:mr_op_nc_all_10}. We also report the Mean Rank and Mean Penalty of embedding methods – on datasets with few labels – in Figure \ref{fig:mr_op_nc_10_fc}. The observations we reported with train:test 50:50 split also seem to hold with train:test 10:90 split. Specifically, we observe NetMF is the most competitive method for node classification while Laplacian Eigenmaps method outperforms multiple existing methods on multiple datasets (Blogcatalog, Co-author datasets).  Embedding methods such as DeepWalk and LINE also perform well on most datasets.

\end{itemize}

%% file: conclusion.tex
Network representational learning has attracted lot of attention in past few years. 
An interested reader can refer to the survey of network embedding methods  \cite{cai2018comprehensive,  hamilton2017representation, zhang2018network}. The surveys focus on categorization of the embedding methods based either encoder-decoder framework \cite{hamilton2017representation} or novel taxonomy \cite{cai2018comprehensive, zhang2018network} but does not provide experimental comparison of the embedding methods. 
There does exist one other experimental survey of network embedding methods \cite{goyal2018graph}. However there are key differences. First, we present a systematic study on a larger set of embedding methods, including several more recent ideas, and on many more datasets (15 vs 7). Specifically, we evaluate 12 embedding methods + 2 efficient heuristics on 15 datasets.  Second, there are several key differences in terms of results reported and reproducability. In our work we carefully tune all hyperparameters of each method as well as the logistic classifier (and include information in our reproducability notes).
As a concrete example of where such careful tuning can make a difference consider that on Blogcatalog with a train-test split of 50:50, Goyal {\it et al}, achieve Macro-f1 score of 3.9\% while with tuning the hyper-parameters of logistic regression we achieve a Macro-f1 score of 29.2\%. Third, our analysis reveals several important insights on the role of context, role of different link prediction evaluation strategies (dot product vs classifier), impact of non-linear classifiers and many others. All of these provide useful insights for end-users as well as guidance for future research and evaluation in network representational learning and downstream applications.  Fourth, we also provide a comparison against simple but effective task-specific baseline heuristics which will serve as useful strawman methods for future work in these areas.

To conclude, we identify several issues in the current literature: lack of standard assessment protocol, use of default parameters for baselines, lack of standard benchmark, ignorance of task-specific baselines. Additionally, we make the following observations:

\begin{itemize}[noitemsep,leftmargin=*]
    \item MNMF and  NetMF are the most effective baselines for the link prediction and node classification task respectively.  
    \item  No one method completely outperform the other methods on both  link prediction and node classification tasks. 
    \item If one considers Laplacian Eigenmaps as a baseline, the classifier parameters should be tuned appropriately.
    \item The Link Prediction Heuristic we present is simple, efficient to compute and offers competitive performance. The Node Classificaton Heuristic is also simple and efficient to compute and is effective  on data-sets with fewer labels.
    \item For both tasks, some methods are impervious to the use of context whereas for other methods context helps significantly.
    \item While comparing embeddings methods through link prediction task, the superiority of the embedding methods should be asserted by leveraging the classifier.
\end{itemize}


We hope the insights put forward in this study are helpful to the community and encourage the comparison of novel embedding methods with the task-specific competitive methods and proposed task-specific heuristics.

%% file: tables.tex
\begin{table*}[bt]

\caption{ Link Prediction performance measured with AUROC. The ``-" represents that the method did not scale on the particular dataset.}
\resizebox{1.0\linewidth}{!}{
\label{table:lp_auroc}
\input{table_link_prediction_aucroc.tex}
}

\vspace{1em}

\caption{ Link Prediction performance measured with AUPR. The ``-" represents that the method did not scale on the particular dataset.}
\resizebox{1.0\linewidth}{!}{
\label{table:lp_aupr}
\input{table_link_prediction_aupr.tex}
}
\vspace{1em}
\caption{Node Classification performance measured in terms of Micro-f1 with train-test split of 50:50  with Logistic Regression. The ``-" represents that the method did not scale on the particular dataset.}
\resizebox{1.0\linewidth}{!}{
\input{table_node_classi_mif1.tex}
}

\vspace{1em}

\caption{Node Classification performance measured in terms of  Macro-f1 with train-test split of 50:50 with Logistic Regression. The ``-" represents that the method did not scale on the particular dataset.}
\resizebox{1.0\linewidth}{!}{
\input{table_node_classi_maf1.tex}
}

\vspace{1em}

\end{table*}

%% file: table_link_prediction_aucroc.tex
\begin{tabular}{lccccccccccccc}
\toprule
\toprule
     Datasets &  Heuristics &  LapEig &  DeepWalk &  Node2Vec &  Verse &  LINE  &  GraRep &  HOPE &  SDNE &  NetMF &  MNMF &   VAG &   WYS \\
\midrule
\midrule
 W-Texas & 77.7 & 73.1 & 79.7 & 83 & 81.7 & 78.2 & 78.7 & 78.7 & 82.4 & 80.9 &  \textbf{96.0} & 78.6 &  83.0 \\
W-Cornell &  81.5 & 77.3 & 79.2 & 82 & 87 & 77.5 & 84.4 & 79.9 & 80.2 & 81.2 &  \textbf{96.7} & 74.8 &  84.4 \\
W-Washington & 75.3 & 70.1 & 75.3 & 75 & 82.2 & 72.9 & 78.1 & 72.8 & 76.7 & 75.5 &  \textbf{97.5} & 73.9 &  79.2 \\
W-Wisconsin &  79.4 & 71.4 & 80.7 & 78 & 88.2 & 72.5 & 84.3 & 75.5 & 76.9 & 82.3 &  \textbf{98.9} & 73.9 &  84.5 \\
PPI &  90.9 & 78.2 & 89.1 & 88.3 & 89.6 & 87.8 & 90 & 88.4 & 89.3 & 87.3 &  \textbf{96.9} & 87.4 &  91.5 \\
Wikipedia &  91.6 & 77.9 & 90.9 & 90.9 & 91.3 & 91.2 &  \textbf{92.3} & 90.4 & 50 & 91.4 & 88.4 & 89.5 &  \textbf{92.3} \\
Wiki-Vote &  91.5 & 83.5 & 97.4 & 97.6 & 94.9 & 96.6 & 88.4 & 97.8 & 96.6 & 95.5 & 92.2 & 94.3 &  \textbf{98.2} \\
BlogCatalog &  95.2 & 77.4 & 94.3 & 95 &  \textbf{97.3} & 95.2 & 96.2 & 95.3 & 95.6 & 95.1 & 94 & 94.8 &  96.0 \\
DBLP (Co-Author) & 95.6 & 93.3 & 96 & 95.4 & 97.9 & 94.3 & 97.1 & 89.6 & 50 & 95.9 &  \textbf{99.4} & 94.1 &  96.8 \\
Pubmed & 87.7 & 89.6 & 89.1 & 89.3 & 96.6 & 92.7 & 77.7 & 90.1 & 88.9 & 89.8 & 94.3 & 93.6 &  \textbf{97.0} \\
CoCit (microsoft) &  89.5 & 95.6 & 97.6 & 97.3 & 83.7 & 97.2 &  \textbf{97.9} & 94.5 & 91.9 & 96.9 & 96.6 & 96.2 &  - \\
P2P &  83.8 & 69.9 & 88.2 & 88.3 & 77.6 & 91.2 & 71.8 & 88.6 & 83.9 & 87.5 &  \textbf{92.3} &  - &  - \\
Flickr & 92.4 & 93 & 95.8 & 94.7 & 72.6 & 95.2 & 95.5 & 96.5 & 93 &  \textbf{97.2} &  - &  - &  - \\
Epinions & 92.2 & 90.9 & 93.3 & 93.4 & 91.9 & 91.6 &  \textbf{93.7} & 92.7 & 92.7 & 92.8 &  - &  - &  - \\
Youtube &  96.2 & 96 & 93.6 & 91.4 &  \textbf{97.6} & 96.5 & 91.4 & 92.4 &  - &  - &  - &  - &  - \\
\bottomrule

\end{tabular}

%% file: table_link_prediction_aupr.tex
\begin{tabular}{lccccccccccccc}
\toprule
\toprule
     Datasets &  Heuristics &  LapEig &  DeepWalk &  Node2Vec &  Verse &  LINE &  GraRep &  HOPE & SDNE &  NetMF &    MNMF &   VAG &   WYS \\
\midrule
\midrule
W-Texas &  81.8 & 78 & 81.9 & 85 & 85 & 82.2 & 82.1 & 82.9 & 85.5 & 83.6 &  \textbf{96.0} & 80.6 &  85.2 \\
W-Cornell &  81.9 & 78.9 & 79.8 & 81 & 87.8 & 76.7 & 86.2 & 79.8 & 79.5 & 82 &  \textbf{96.7} & 75.6 &  86.6 \\
W-Washington & 80.3 & 75 & 76.5 & 78 & 86.5 & 75.5 & 82.9 & 77.2 & 81.5 & 80.4 &  \textbf{97.7} & 78.7 &  83.4 \\
W-Wisconsin &  82.3 & 74.8 & 81.6 & 79 & 90.7 & 76.4 & 87.4 & 78.8 & 80.7 & 85.2 &  \textbf{98.5} & 78.1 &  86.7 \\
PPI &  91.4 & 80.7 & 90.4 & 89.5 & 90.7 & 88.1 & 90.8 & 89.2 & 90.2 & 87.9 &  \textbf{96.5} & 88.1 &  92.2 \\
Wikipedia &  93 & 76 & 92.5 & 92.3 & 92.8 & 92.8 & 93.1 & 91.8 & 75 & 92.8 & 89.9 & 91.4 &  \textbf{93.5} \\
Wiki-Vote &  87.9 & 82.1 & 96.9 & 97.2 & 94.8 & 95.3 & 84 & 96.8 & 96.3 & 93.8 & 87.6 & 94.3 &  \textbf{97.4} \\
BlogCatalog &  95.1 & 77.5 & 94.3 & 94.8 &  \textbf{97.9} & 95.1 & 96 & 95 & 95.5 & 94.8 & 93.6 & 94.6 &  96.1 \\
DBLP (Co-author) & 96.7 & 93.8 & 96.8 & 96.1 & 98.2 & 95.6 & 97.4 & 90.9 & 75 & 96.7 &  \textbf{99.2} & 95.2 &  97.3 \\
Pubmed & 85 & 85 & 81.5 & 82.3 & 96.8 & 90.3 & 74.1 & 91.4 & 90.5 & 86.1 & 90.3 & 95.2 &  \textbf{96.9} \\
Cocit (microsoft) &  91.9 & 95.5 &  \textbf{97.9} & 97.5 & 76.4 & 97.7 &  \textbf{97.9} & 95.3 & 93.5 & 97.1 & 95.2 & 96.4 &  - \\
P2P &  79.3 & 68.1 & 84.3 & 84.5 & 68.3 & 88.6 & 71.3 & 85.8 & 80.8 & 84 &  \textbf{89.6} &  - &  - \\
Flickr & 92.5 & 95 & 96.1 & 95 & 70.9 & 95.5 & 95.7 & 96.7 & 94 &  \textbf{97.6} &  - &  - &  - \\
Epinions & 89.2 & 89.5 & 91.7 & 91.9 & 88.6 & 88.8 &  \textbf{93.0} & 91.5 & 91.6 & 91.8 &  - &  - &  - \\
Youtube &  96.7 & 96.7 & 95 & 93 &  \textbf{98.2} & 97 & 92.2 & 94 &  - &  - &  - &  - &  - \\
\bottomrule
\end{tabular}

%% file: table_node_classi_mif1.tex
\label{table:nc-micro}
\begin{tabular}{lccccccccccccc}
\toprule
\toprule
           Datasets &  Heuristics &  LapEig &  DeepWalk &  Node2Vec &  Verse &  LINE &  GraRep &  HOPE &  SDNE &  NetMF & MNMF &   VAG &   WYS \\
\midrule
\midrule
W - Texas &       61.8 &   54.6 &     55.1 &     57.2 &  54.5 &  61.8 &   56.1 &  59.1 &  58.0 &  67.1 &  58.0 &  54.8 &  60.6 \\
    W - Cornell &       42.1 &   30.9 &     40.5 &     34.3 &  35.4 &  44.1 &   40.9 &  41.9 &  48.5 &  48.1 &  36.1 &  40.3 &  41.8 \\
 W - Washington &       65.0 &   43.3 &     56.0 &     58.4 &  51.5 &  65.3 &   46.7 &  62.8 &  60.5 &  61.1 &  60.4 &  59.0 &  65.3 \\
  W - Wisconsin &       51.7 &   41.5 &     52.3 &     45.6 &  41.7 &  52.9 &   50.8 &  51.5 &  51.3 &  56.7 &  53.9 &  48.1 &  53.2 \\
                PPI &       10.8 &   22.3 &     21.4 &     21.0 &  19.7 &  19.9 &   20.4 &  18.8 &  17.4 &  21.3 &  18.6 &  19.2 &  22.6 \\
          Wikipedia &       41.9 &   46.3 &     50.0 &     51.4 &  43.8 &  56.3 &   58.8 &  57.9 &  52.4 &  58.4 &  48.1 &  41.1 &  44.4 \\
        Blogcatalog &       17.1 &   42.1 &     41.5 &     41.7 &  35.5 &  38.6 &   41.3 &  34.4 &  29.5 &  41.7 &  21.6 &  17.1 &  38.9 \\
   DBLP (Co-Author) &       37.3 &   37.1 &     35.9 &     35.6 &  37.2 &  37.0 &   35.7 &  36.0 &  37.4 &  36.6 &  36.2 &  36.2 &   - \\
             Pubmed &       57.8 &   81.7 &     81.5 &     81.1 &  63.0 &  64.4 &   79.1 &  74.7 &  67.7 &  80.0 &  77.1 &  63.5 &  73.6 \\
  CoCit (microsoft) &       25.0 &   43.0 &     46.3 &     46.7 &  46.0 &  46.5 &   46.6 &  44.6 &  38.1 &  43.5 &  44.7 &  43.5 &   - \\
             Flickr &       19.1 &   34.0 &     35.6 &     35.1 &  30.1 &  33.4 &   10.5 &  28.7 &  30.8 &  34.2 &   - &   - &   - \\
            Youtube &       24.4 &    - &     40.7 &     40.3 &  38.5 &  40.3 &   38.0 &  38.7 &   - &   - &   - &   - &   - \\
\bottomrule
\end{tabular}

%% file: table_node_classi_maf1.tex
\label{table:nc-macro}
\begin{tabular}{lccccccccccccc}
\toprule
\toprule
           Datasets & Heuristics & LapEig & DeepWalk & Node2Vec & Verse &  LINE &   GraRep &  HOPE & SDNE & NetMF & MNMF &   VAG &   WYS \\
\midrule
\midrule
      W -Texas &       42.1 &   18.1 &     26.9 &     22.7 &  18.1 &  36.4 &   25.0 &  34.4 &  39.5 &  49.9 &  23.8 &  27.8 &  40.3 \\
    W -Cornell &       22.2 &   21.3 &     28.1 &     23.2 &  22.4 &  25.0 &   27.7 &  28.7 &  13.1 &  32.8 &  23.5 &  20.6 &  26.5 \\
 W -Washington &       32.2 &   22.2 &     24.3 &     27.3 &  23.1 &  30.6 &   28.7 &  30.2 &  29.1 &  29.7 &  28.6 &  26.3 &  31.1 \\
  W -Wisconsin &       24.7 &   29.0 &     31.9 &     23.9 &  21.9 &  27.8 &   34.7 &  28.3 &  26.1 &  34.8 &  33.8 &  25.5 &  33.9 \\
                PPI &        6.0 &   17.9 &     18.1 &     18.0 &  16.5 &  16.9 &   17.4 &  15.9 &  15.2 &  17.5 &  15.9 &  13.1 &  17.9 \\
          Wikipedia &        5.5 &   10.4 &     11.9 &     12.9 &   8.2 &  18.2 &   18.3 &  20.1 &  14.1 &  18.4 &  11.0 &   3.8 &  10.1 \\
        Blogcatalog &        3.1 &   29.2 &     27.3 &     27.9 &  22.1 &  23.6 &   28.9 &  20.8 &  14.8 &  28.8 &   8.2 &   3.1 &  26.3 \\
   DBLP (Co-Author) &       18.1 &   20.1 &     30.0 &     29.4 &  20.6 &  19.2 &   30.5 &  28.6 &  21.1 &  30.0 &  26.6 &  27.8 &   - \\
             Pubmed &       48.9 &   80.2 &     80.1 &     79.8 &  58.0 &  61.3 &   77.6 &  73.0 &  63.3 &  78.4 &  75.4 &  60.6 &  71.6 \\
  CoCit (microsoft) &       12.6 &   27.3 &     34.3 &     34.2 &  33.3 &  33.8 &   34.8 &  32.8 &  27.8 &  34.0 &  30.4 &  29.2 &   - \\
             Flickr &        1.7 &   20.4 &     21.2 &     20.7 &  17.6 &  18.2 &    0.9 &  11.4 &  14.9 &  20.2 &   - &   - &   - \\
            Youtube &        9.3 &    - &     34.7 &     34.0 &  32.1 &  33.1 &   30.0 &  30.8 &   - &   - &   - &   - &   - \\
\bottomrule

\end{tabular}